\documentclass[preprint,3p,times]{elsarticle}
\usepackage{graphicx,import}
\usepackage{amssymb,amsfonts,amsmath}
\usepackage[utf8]{inputenc}
\usepackage{subcaption}
\usepackage{mathtools}
\usepackage{epsfig}
\usepackage{epstopdf}
\usepackage{subeqnarray}
\usepackage{array}
\usepackage{cancel}
\usepackage{kbordermatrix}
\usepackage{blkarray}
\usepackage{float}
\usepackage{amsmath}

\usepackage{xcolor}

\floatstyle{plain}
\newfloat{TwoColsEq}{h}{lop}[section]

\usepackage[colorlinks,bookmarksopen,bookmarksnumbered,citecolor=red,urlcolor=red]{hyperref}

\journal{..}

\providecommand{\FIG}[1]{Figure~\ref{#1}}
\providecommand{\EQ}[1]{Eq.~(\ref{#1})}

\providecommand{\HAT}[1]{#1} % Without hats

\providecommand{\x}[0]{\mathbf{x}}
\providecommand{\dx}[0]{\dot{\mathbf{x}}}

\providecommand{\e}[0]{\mathbf{e}}

\providecommand{\M}[0]{{\mathbf{M}}}

\providecommand{\oM}[0]{\overline{\mathbf{M}}}

\providecommand{\Q}[0]{{\mathbf{Q}}}

\providecommand{\oQ}[0]{\overline{\mathbf{Q}}}

\providecommand{\R}[0]{{\mathbf{R}}}

\providecommand{\B}[0]{{\mathbf{B}}}

\providecommand{\PHI}[0]{{\boldsymbol{\Phi}}}

\providecommand{\LAMBDA}[0]{{\boldsymbol{\Lambda}}}

\providecommand{\F}[0]{{\mathbf{Q}}}

\providecommand{\OBS}[0]{{\mathbf{o}}}

\providecommand{\q}[0]{{\mathbf{q}}}
\providecommand{\dq}[0]{{\dot{\mathbf{q}}}}
\providecommand{\ddq}[0]{{\ddot{\mathbf{q}}}}

\providecommand{\z}[0]{{\mathbf{z}}}
\providecommand{\dz}[0]{{\dot{\mathbf{z}}}}
\providecommand{\ddz}[0]{{\ddot{\mathbf{z}}}}

\providecommand{\lp}[0]{\left(}
\providecommand{\rp}[0]{\right)}

\providecommand{\Phiq}[0]{\mathbf{\Phi_q}}
\providecommand{\Phiqq}[0]{\mathbf{\Phi_{\q\q}}}
\providecommand{\dotPhiqq}[0]{(\dot{\PHI}_{\q})_{\q}}
\providecommand{\Ga}[0]{\mathbf{\Gamma}}

\providecommand{\inv}[0]{{\hspace*{0.1cm} -1}}

\newcommand{\norm}[1]{\left\lVert#1\right\rVert}

\providecommand{\Sp}[0]{\hspace*{0.2cm}}
\providecommand{\Spp}[0]{\hspace*{0.3cm}}

\begin{document}

\begin{frontmatter}

% -------------- frontmatter COMMANDS -------------------

% \title{A factor graph approach to generic identification algorithms in Multibody dynamics}

%\title{A General Framework for Tracking, Parameter Identification, and Force Estimation in Multibody Dynamics using Factor Graphs}

\title{A general framework for modeling and dynamic simulation\\ of multibody systems using factor graphs}
\author[cor1,add1]{José-Luis Blanco-Claraco\corref{cor1}}
\ead{jlblanco@ual.es}
\ead[url]{https://w3.ual.es/personal/jlblanco/}
\author[add2]{Antonio Leanza}
\author[add3]{Giulio Reina}

\cortext[cor1]{Corresponding author:}
\address[add1]{Department of Engineering, University of Almería, CIESOL. Campus de Excelencia Internacional Agroalimentario, ceiA3, 04120 Almería, Spain}
\address[add2]{Department of Innovation Engineering, University of Salento, via Monteroni, 73100, Lecce, Italy}
\address[add3]{Department of Mechanics, Mathematics, and Management, Polytechnic of Bari, via Orabona 4, 70126 Bari, Italy}

\begin{abstract}
In this paper, we present a novel general framework grounded in the factor graph theory to solve kinematic and dynamic problems for multi-body systems.
Although the motion of multi-body systems is considered to be a well-studied problem and various methods have been proposed for its solution, a unified approach providing an intuitive interpretation is still pursued.
We describe how to build factor graphs to model and simulate multibody systems using both, independent and dependent coordinates.
Then, batch optimization or a fixed-lag-smoother can be applied to solve the underlying optimization problem that results in a highly-sparse nonlinear minimization problem.
The proposed framework has been tested in extensive simulations and validated against a commercial multibody software. We release a reference implementation as an open-source C++ library, based on the GTSAM framework, a well-known estimation library.
Simulations of forward and inverse dynamics are presented, showing comparable accuracy with classical approaches. The proposed factor graph-based framework has the potential to be integrated into applications related with motion estimation and parameter identification of complex mechanical systems, ranging from mechanisms to vehicles, or robot manipulators.
\end{abstract}


\end{frontmatter}

\section*{List of symbols}
%\addcontentsline{toc}{chapter}{List of Symbols}

\begin{tabular}{cp{0.9\textwidth}}
%$\hat{\mathbf{i}},\hat{\mathbf{j}},\hat{\mathbf{k}}$ & Orthonormal unit vectors, or \emph{versors}. \\
$\q(t), \dq(t), \ddq(t)$ & Vector of generalized dependent coordinates, velocities, and accelerations, respectively. \\
$\z(t), \dz(t), \ddz(t)$ & Vector of generalized independent coordinates, velocities, and accelerations, respectively. \\
$\PHI\left(\q,t\right)$ & Vector of constraint equations. \\
$\Phiq$ & Jacobian matrix of the constraints $\PHI$. \\
$\PHI_t$ & Partial derivatives of constraints $\PHI$ with respect to time $t$. \\
$\Phiqq, \dotPhiqq$ & Jacobians of $\Phiq$ and $\dot{\PHI}_{\q}$ with respect to $\q$ (third-order tensors). \\
$\mathbf{b}$ & Constraint velocities. \\
$\mathbf{c}$ & Constraint acceleration. \\
$\boldsymbol{\lambda}$ & Vector of Lagrangian multipliers. \\
$C$ & Active "branch" or "configuration" of a mechanism. \\
%$\OBS_t$ & Vector of observations (sensors) for time step $t$. \\
$\Q_t$ & Generalized forces for time step $t$. \\
$\M$ & Mass matrix for a mechanism (dependent coordinates).  \\
$\oQ$ & Reduced generalized forces vector. \\
$\oM$ & Reduced mass matrix for a mechanism (independent coordinates). \\
$\R$ & The $\R$ matrix, mapping increments between independent and dependent velocities.\\
%$\B$ & Rows of constant projection matrix mapping changes between dependent and independent velocities. \\
$n$ & Number of dependent coordinates in $\q(t)$. \\
$m$ & Number of scalar constrain equations in $\PHI$. \\
$\mathbb{R}$ & The set of Real numbers. \\
\end{tabular}

% ======================= INTRO ========================
\section{Introduction}
\label{sect:intro}

% Aim of the paper:

% Brief related works, estimation in mechanisms:
Dynamic simulation of multibody systems mainly refer to inverse and forward problems.
Inverse dynamics deals with the determination of the driving forces that generate a given motion, as well as the constraint reaction forces. The solution to the inverse dynamics problem can be obtained for example using the Newton-Euler (N-E) method \cite{SIC}, that results in efficient recursive algorithms, e.g., Rigid Body Dynamics Library (RBDL) \cite{RBDL} and similar methods \cite{ORI}. Conversely, forward dynamics involves the motion estimation of a multi-body system over time under the given applied forces and initial conditions. Therefore, in a direct dynamic problem, the motion is the result of the force system that generates it. From a mathematical perspective, forward dynamics is computationally intensive as it entails the integration of a system of nonlinear ordinary differential equations. The most common formulations that deal with forward dynamic are the Composite-Rigid-Body Algorithm (CRBA) \cite{walker_efficient_1982} and the Articulated-Body Algorithm (ABA) \cite{featherstone_robot_2000}.
However, the above mentioned methods are not suited for systems with closed-loops like parallel robots, for which  more complex and expensive procedures are required to solve both inverse and forward dynamics, as described for example in \cite{WAN}.\\ Therefore, a single algorithm that can solve all types of dynamics problems has not been fully established. One notable effort is the work by Rodriguez \cite{ROD}, who proposed a unified approach based on Kalman filtering and on the concept of smoothing to solve dynamics problems. This approach has been further extended to solve the forward dynamic problem of closed kinematic chains \cite{ROD1}. Another effort in defining a unifying framework has been given by \cite{JAI}, who analyzed various algorithms for serial chain dynamics. In \cite{ASC}, an attempt is proposed to unify the CRBA and ABA derivation as two elimination methods to solve forward dynamics.\\
To the best of our knowledge, factor graphs have not been applied to solve the general multibody kinematics and dynamics in the existing literature.
The closest works are the preprints \cite{xie_batch_2020,xie_unified_2020},
where factor graphs are indeed employed to solve kinematic and dynamic problems, although their proposed graph structure is applicable to open-loop robot arms only.

The present work advocates factor graphs as a unifying graphical language in which to express the classical kinematics and dynamics algorithms, with the additional potential to develop novel and advanced state estimators.
The main contributions of this work are:
\begin{itemize}
  \item a factor graph-based representation of dynamics problems, which is a insightful visualization of the underlying sparse constraints between all involved variables,
  \item a unified method, which can solve inverse and forward dynamics for either open or closed kinematic chains,
  \item a flexible framework that can be expressed and solved for both dependent and independent coordinates.
\end{itemize}

Also, note that despite the implementation described in this manuscript is focused on planar mechanisms, it is perfectly suitable to spatial systems without changes at the level of factor graphs.
Additionally, our approach allows more powerful and flexible schemes for state and parameter estimation to be implemented in contrast with standard methods based on Kalman filtering \cite{cuadrado2009real, pastorino2013ekf, naets2014online, sanjurjo2017accuracy, REI}.
However, such applications are left for future extensions of this work to keep the present manuscript focused on the
key ideas on how to apply factor graphs to multibody motion problems.

The proposed approach draws on the formalism of \emph{graphical models}, a powerful tool borrowing concepts from statistics and graph theory \cite{loeliger2004ifg,bishop2006pattern}.
By addressing the multibody simulation problem from the perspective of the \emph{variable structure}, graphical models allow us creating efficient
estimators for any combination of observed and hidden variables,
effectively unifying the problems of kinematic and forward dynamic analysis
(predicting or estimating the trajectory of a MB system),
inverse dynamics, and parameter identification (e.g., inertial properties of the bodies involved, external disturbance forces, friction in the joints, etc.).
All those problems end up to be formulated as a sparse nonlinear cost function
built from a library of reusable ``building blocks'' (the \emph{factors})
on which efficient solvers can then be applied.
The framework is suitable for either offline batch analysis or online real-time operation that represents another clear advantage of the proposed approach.

The rest of the paper is structured as follows.
Sections \ref{sect:primer.mbd} and \ref{sect:bg.graphical.models} first
provide the required background on multibody dynamics and graphical models, respectively.
Next, section~\ref{sect:dbn.mbd} presents a methodology for the application of Bayesian networks to multibody dynamics problems,
whereas section~\ref{sect:fromMBDtoFGdep} particularizes such networks as factor graphs for a number of practical problems.
Individual factors used in those graphs are described in detail in section~\ref{sect:Factors}.
Numerical examples are provided in section~\ref{sect:Experiments}, and we end sketching some conclusions in section~\ref{sect:Conclusions}.

% ================== Background MBD  ====================
\section{Review of Multibody Dynamics} \label{sect:primer.mbd}
% Minimum reminder/summary of MBD theory (with references to "the red book", etc.)

In this section, fundamentals of multibody system motion analysis are briefly recalled, whereas the interested reader can refer for more details to the wide related Literature, e.g. \citep{garcia1994}. A \textit{multibody system} is an assembly of two or more bodies (or \emph{elements}) constrained to each other to fulfill a given motion law. In many practical applications, these elements may be considered rigid and, throughout this paper, we will work under this assumption even though the proposed framework may be further extended to include body flexibility.\\
One of the key decisions to take when modeling a MBS is selecting the set of \emph{generalized coordinates} that will be used to represent it. Using \textit{independent coordinates} $\z$ allows one to deal with the lowest number of parameters, i.e. the number of DOFs of the system.
% JL: Dont get into that discussion here since its an introduction?
%But sometime this system of coordinates may lead to insufficient and/or non-unique system description (\textbf{ADD FIGURE OF OUR 4-BAR WITH CRANK ANGLE NOT SUFFCIENT}).
A second choice is to adopt \textit{dependent coordinates} $\q$, in a number larger than that of DOFs but able to describe all multibody system points univocally.
When dependent coordinates are used, the corresponding set of \textit{constraint equations} must be included as well for a complete system analysis.
There are different kinds of dependent coordinates \citep{garcia1994}: \textit{Relative Coordinates}, \textit{References Point Coordinates}, \textit{Natural Coordinates}, and a combination of the previous ones (\textit{Mixed Coordinates}).
Natural coordinates, mixed with relative coordinates where needed, will be assumed in this work.

%The interested reader can find plenty of details on the multibody analysis concepts used throughout this work in the excellent existing literature \citep{garcia1994,shabana2005dynamics,roberson2012dynamics}.
In the remainder of this Section, the MBS motion equations are developed and expressed in terms of both dependent and independent coordinates.

\subsection{Dependent coordinates formulation}
\label{sect:DepCoordFormulation}

For any given MBS with $f$ dofs, the use of $n$ dependent coordinates expressed by the vector $\q$ requires $m = n-f$ constraint equations, which form the following set of equations
\begin{equation}
\PHI\left(\q(t),t\right)=\mathbf{0}
\label{eq:Constraints}
\end{equation}
Thus, it is assumed that there are at least as many equations as there are unknown coordinates. To solve the kinematic problem, the time derivative of \EQ{eq:Constraints} is required, one time for velocity analysis and two times for acceleration analysis, leading to the following set of equations
\begin{equation}
\Phiq(\q(t),t) ~ \dq = -\PHI_t \equiv \mathbf{b}
\label{eq:ConstraintsVel}
\end{equation}
\begin{equation}
\Phiq(\q(t),t) ~ \ddq = -\dot{\PHI}_t-\dot{\PHI}_{\q}\dq \equiv \mathbf{c}
\label{eq:ConstraintsAcc}
\end{equation}
where $\dot{\mathbf{q}}$ is the vector of dependent velocities, $\Phiq\in\mathbb{R}^{m\times n}$ the Jacobian matrix of \EQ{eq:Constraints} respect to $\q$, and $\PHI_t$ the time derivative of constraint equations that is equal to the null vector for time-independent constraints.
\\ To solve the dynamic problem, external forces and inertia forces need to be considered. From the classical Newton's law, one obtains the following set of differential equations that expresses the force equilibrium equations
\begin{equation}
\M\ddq + \Phiq^\top\boldsymbol{\lambda} = \F
\label{eq:NewtonLaw}
\end{equation}
where $\boldsymbol{\lambda}$ is the vector of Lagrange multipliers, $\M$ is the system mass matrix and vector $\F$ contains the generalized external forces. Since \EQ{eq:NewtonLaw} is a  system of $n$ equations in $n+m$ variables, by adding the $m$ equations (\ref{eq:ConstraintsAcc}) one obtains the following system
\begin{equation}
\begin{bmatrix}
\mathbf{M} && \mathbf{\Phi_q}^\top \\
\mathbf{\Phi_q} && \mathbf{0}
\end{bmatrix}
\left[ \begin{array}{c}
\ddq \\ \boldsymbol{\lambda}
\end{array} \right] =
\left[ \begin{array}{c}
\F \\ \mathbf{c}
\end{array} \right]
\label{eq:qDynSys}
\end{equation}

Normally \cite{garcia1994,shabana2005dynamics}, this equation is solved for the extended vector of unknowns that includes both, $\ddq$ and $\boldsymbol{\lambda}$. In our framework, we are specifically interested in the generalized accelerations $\ddq$. Therefore, by applying the block matrix inversion lemma (see \ref{sect:Appendix.I.BlockMatrix}) to \EQ{eq:qDynSys} and keeping the first row only, it leads to the following \textit{equation of motion} expressed in terms of \textit{dependent coordinates}:

\begin{equation}
\label{eq:ddq}
\ddq = \lp \M^{-1} - \M^{-1}\Phiq^\top \Ga^{-1}\Phiq\M^{-1} \rp \F + \lp \M^{-1}\Phiq^\top \Ga^{-1} \rp \mathbf{c}
\end{equation}

\noindent where we defined the auxiliary term $\Ga \lp \q \rp$ as the $m\times m$ square symmetric matrix

\begin{equation}
\label{eq:def.Gamma}
\Ga = \Phiq\M^{-1}\Phiq^\top,
\end{equation}

\noindent introduced for convenience in subsequent derivations.

%Based on this state-space reduction method known as projection matrix-R method, which allows to convert the DAEs of the MBS model into ordinary differential equations (ODE), the implementation of a batch estimator (? a little vague) is straightforward, and no additional sensitivity equations are needed.

\subsection{Independent coordinates formulation}

Independent coordinates $\z$, ensure the minimum number of variables, i.e. the number of DOFs. However, multibody systems can be more difficult to analyze with respect to dependent ones.
First, the matrix $\R$ is introduced as
\begin{equation}
\dq = \frac{d\q}{dt} = \frac{\partial \q}{\partial \z}\frac{\partial \z}{\partial t} = \mathbf{R}(\q)\dz
\label{eq:dotq=Rdotz}
\end{equation}
%Then, starting from \EQ{eq:ddq},
%the \textit{equation of motion} in terms of \textit{independent coordinates} will be obtained.
where the columns of $\R$ are $f$ linearly independent vectors that constitute a basis of the nullspace of $\Phiq$.
Then, we express the independent velocities $\dz$ as the projection of the dependent velocities $\dq$ on the rows of a constant matrix $\B$ that we assume satisfying the condition of having $f$ rows linearly independent from each other and from the $m$ rows of $\Phiq$
\begin{equation}
\dz = \B\dq
\label{eq:dotz=Bdotq}
\end{equation}
\noindent By combining \EQ{eq:ConstraintsVel} with \EQ{eq:dotz=Bdotq}, one obtains
\begin{equation}
\begin{bmatrix}
\Phiq \\ \B
\end{bmatrix}
\dq =
\begin{bmatrix}
\mathbf{b} \\ \dz
\end{bmatrix}
\end{equation}
\begin{equation}
\dq =
\begin{bmatrix}
\Phiq \\ \B
\end{bmatrix}^{-1}
\begin{bmatrix}
\mathbf{b} \\ \dz
\end{bmatrix}
= \mathbf{S}\mathbf{b} + \R\dz
\label{eq:dq}
\end{equation}
%\noindent where $\left[\Phiq \hspace*{0.25cm} \B \right]^\top \in \mathbb{R}^{n \times n}$ is a square matrix.
%\textcolor{red}{JL for Antonio: R is not constant??!! It depends on the mechanism position q, right??}
%\textcolor{green}{Yes, it depends by $\q$ in Red Book p. 97 below equation (3.22)}
The columns of matrix $\R$ are the partial velocities with respect to the generalized coordinates $\z$, and the term $\mathbf{Sb}$ represents the partial velocities with respect to time.
%Since $\mathbf{R} = \mathbf{R}\lp\q\rp$
%\begin{equation}
%\dot{\mathbf{R}} = \frac{\partial \mathbf{R}}{\partial \q} \frac{\partial \q}{\partial t} =
% \frac{\partial \mathbf{R}}{\partial \q}\dq
%\end{equation}
%where in general $\partial \mathbf{R} / \partial \q$ is a tensor.

By differentiating \EQ{eq:dotz=Bdotq} and by taking into account \EQ{eq:ConstraintsAcc}, one gets
\begin{equation}
\begin{bmatrix}
\mathbf{c} \\ \ddz
\end{bmatrix} =
\begin{bmatrix}
\Phiq \\ \B
\end{bmatrix} \ddq
\Rightarrow
\ddq =
\left[\mathbf{S} \Spp \R \right]
\begin{bmatrix}
\mathbf{c} \\ \ddz
\end{bmatrix} =
\mathbf{S}\mathbf{c} + \R\ddz
\label{Eq:ddq_ddz}
\end{equation}
From \EQ{eq:NewtonLaw}, pre-multiplied by $\R^\top$
\begin{equation}
\R^\top\M\ddq = \R^\top\F
\end{equation}
and by substituting $\ddq$ in \EQ{Eq:ddq_ddz}
\begin{equation}
\R^\top\M\R\ddz = \R^\top \lp\F-\M\mathbf{Sc} \rp
\end{equation}
Introducing the reduced mass matrix $\oM = \R^\top\M\R$  and force vector $\oQ = \R^\top \lp \F-\M\mathbf{Sc} \rp$, the following equation of motion in terms of independent coordinates can be finally obtained
\begin{equation}
\label{eq:zDynSys}
\ddz = \oM^\inv\oQ
\end{equation}

% ================== Background FG  ====================
\section{Background on graphical models}
\label{sect:bg.graphical.models}

% BASIC INTRODUCTION
A factor graph is a particular type of probabilistic \emph{graphical model} that can be used to describe the structure of an estimation problem \cite{koller2009pgm}.
Factor graphs have found applications in many fields, for example in robot perception \cite{dellaert2017factor}, information theory \cite{worthen2001unified}, signal processing \cite{loeliger2007factor}, and in other areas of robotics, mostly SLAM \cite{grisetti2010tutorial} and computer vision \cite{Forster-RSS-15}, state estimation in legged robots \cite{hartley2018legged}, and kinematic motion planning \cite{sugiarto2013discrete}.

\subsection{Dynamic Bayesian Networks}
% Explain what a graphical model is:

It is instructive to start our discussion by considering a Dynamic Bayesian Network (DBN), a type of graphical model where variables are represented as nodes and \emph{directed} edges stand for \emph{causal} relationships \cite{jensen1996ibn,koller2009pgm}.
The existence of directed edges in a DBN allows us to encode
an expert's knowledge in \emph{causality} relationships between all involved variables in the graph.

The rules to convert a DBN into the kind of graphs we are actually using in this work, factor graphs (FGs), are discussed in section~\ref{sect:fromMBDtoFGdep}.
A relevant point regarding such conversion is that one single DBN can be mapped into different FGs depending on
which subset of the original DBN variables are \emph{known data} and which \emph{unknowns} are required to be found. Let us remark that whereas a DBN displays \emph{all} variables as nodes, in a FG
only \emph{unknown variables} are variable nodes.

\subsection{Factor graphs}

% Factor graphs

Factor graphs (FGs) are bipartite graphs comprising two kinds of nodes: variable nodes, and factor nodes.
Variables are the unknown data to be estimated, and
factors represent any constraint (a \emph{cost function} to be minimized) or relationship between variables.
A crucial aspect of a FG is its \emph{sparsity}: each factor node is only connected to the variables that appear in its cost function. Sparse graph optimization algorithms exist with computational cost almost linear with the number of edges in a FG, as opposed to, for example, the cubic cost of a naive implementation of Kalman Filtering \cite{raitoharju2019computational}. Incremental (e.g. \cite{kaess2012isam2}) and sliding-window (e.g. \cite{strasdat2011double}) approaches exist as well, enabling the efficient estimation of problems with thousands to millions of variables \cite{leutenegger2015keyframe,qin2018vins,agarwal2011building}.

Once a FG is formulated for a given problem, estimating the most-likely value of
all the unknowns becomes a numerical nonlinear minimization problem,
for which very efficient algorithms exist.
As a probabilistic estimator, these optimal values can be assigned an uncertainty.
In general, retrieving the covariance $\boldsymbol{\Sigma}$ of the estimated variables
involves inverting the Hessian of the linearized problem evaluated at the optimal
solution, such that $\boldsymbol{\Sigma} = (\mathbf{J}^\top \LAMBDA \mathbf{J})^{-1} $
with $\mathbf{J}$ the sparse Jacobian of all constraints (factors),
and $\LAMBDA$ the weight matrix representing our level of certainty about each constraint.
The matrix inversion operation is, in general, very costly since it is cubic with the problem dimension.
However, optimized methods exist for the kind of sparse problem at hand, see for example,
\cite[\S B.4]{triggs2000bundle} or \cite{kaess2012isam2}.

% ================== Background FG  ====================
\section{DBN for multibody dynamics}
\label{sect:dbn.mbd}

\begin{figure}
	\centering
	\begin{subfigure}[t]{0.49\textwidth}
		\centering
		\includegraphics[width=1\linewidth]{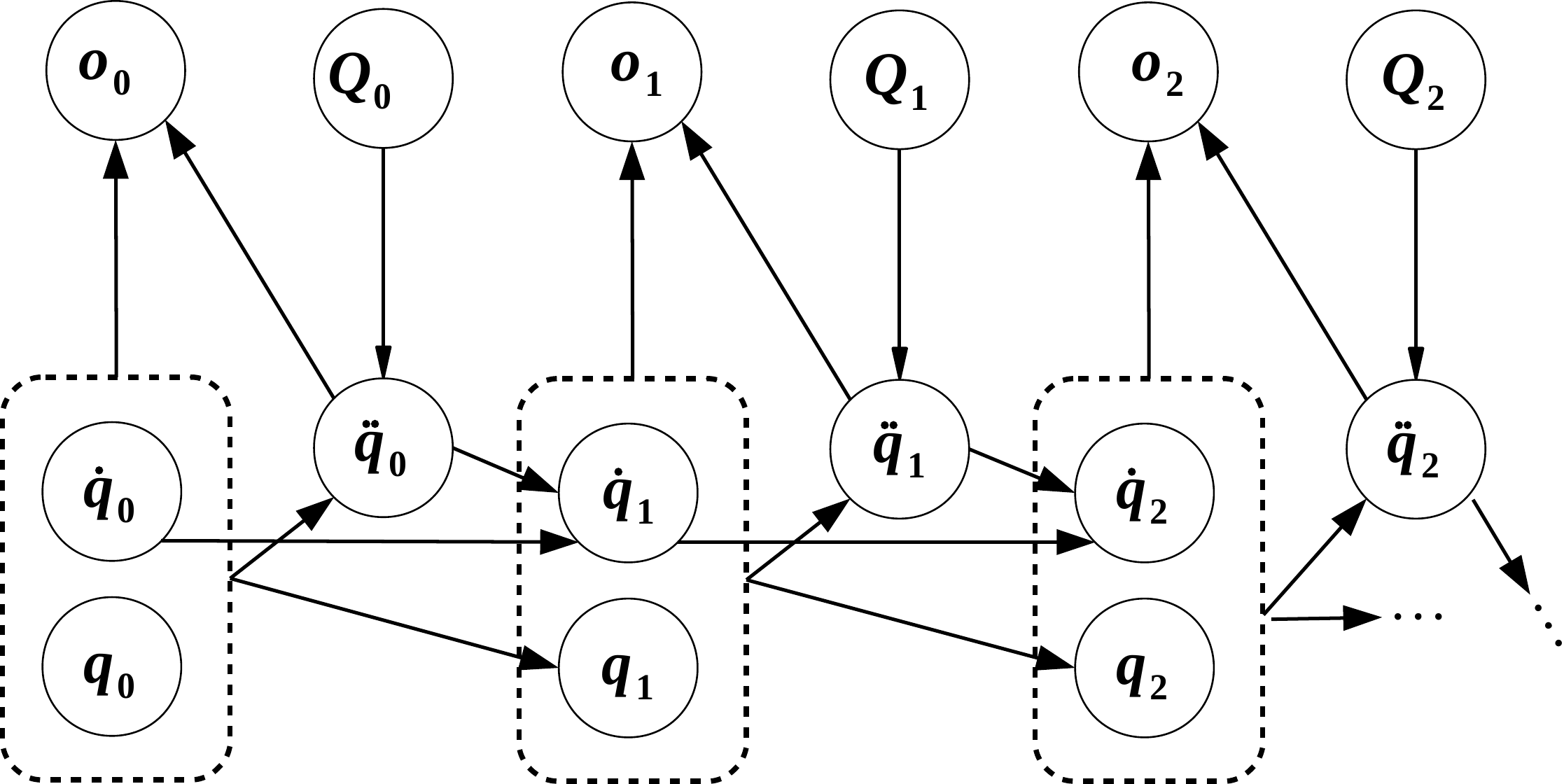}
		\caption{Model for dependent coordinates formulation}
	\end{subfigure}
	\begin{subfigure}[t]{0.49\textwidth}
		\centering
		\includegraphics[width=1\linewidth]{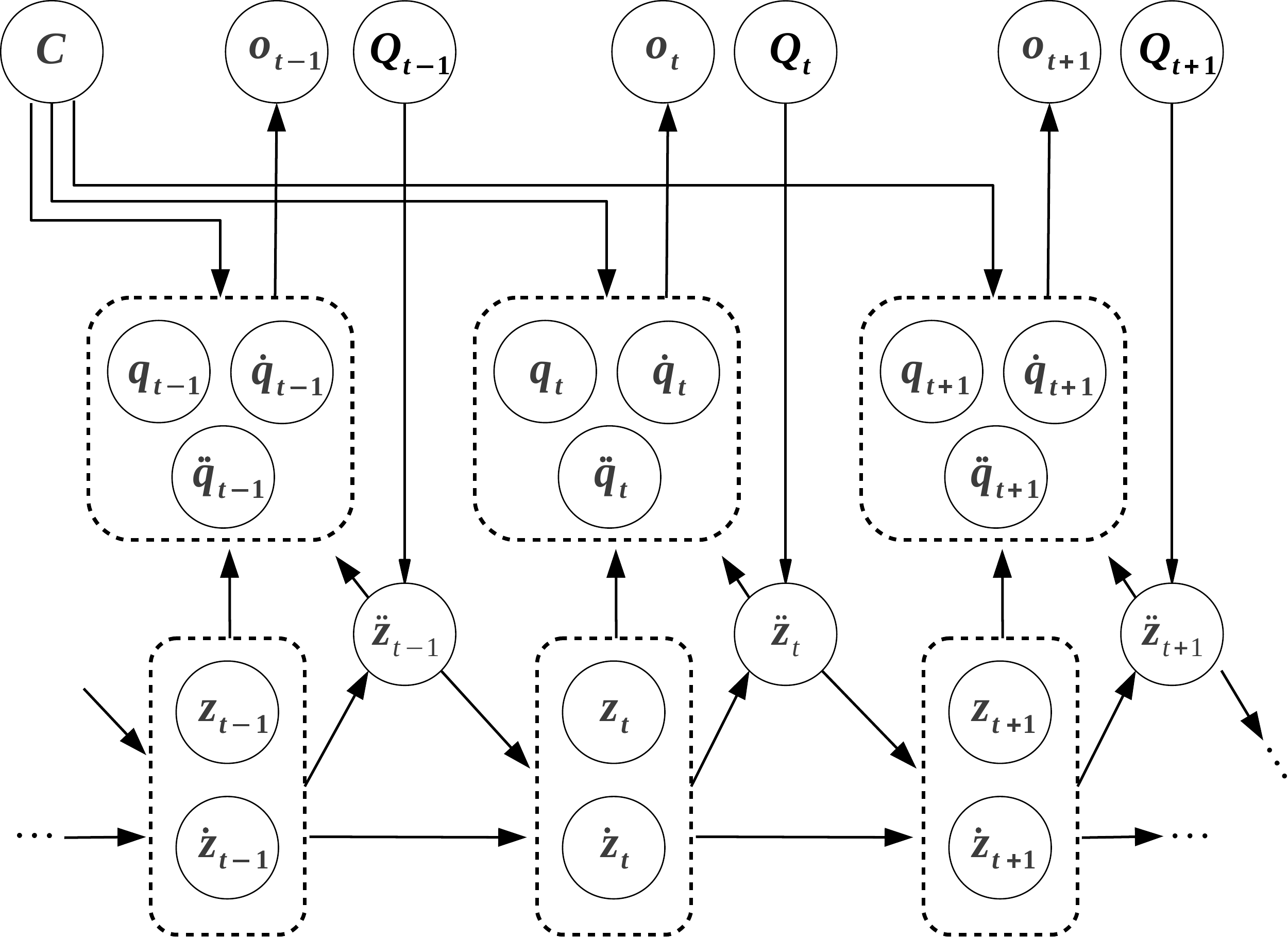}
		\caption{Model for independent coordinates formulation}
	\end{subfigure}
	\caption{
		Dynamic Bayesian Network (DBN) models representing the causality relationship between the variables involved in general multibody dynamics problems over discrete time steps $t$.
		Notation: $\q,\dq,\ddq$ are the dependent coordinates, $\z,\dz,\ddz$ are independent coordinates, $\mathbf{o}$ are sensor observations, $C$ is the active branch of the mechanism, and $\F$ the external forces.
		Dashed boxes represent groups of variables for which input or output directed edges
		affect all members.
		See section~\ref{sect:bg.graphical.models} for a discussion.
	}
	\label{fig:dbn_org}
\end{figure}

% Non linear optimization (why Levenberg-Marquardt)
Since DBNs allow a human expert to specify the \emph{causality} relationships between variables,
we propose the two DBNs in Figure \ref{fig:dbn_org}
as the underlying structure of
a generic multi-body system that is the basis for our proposed framework.
Note the use of \emph{discrete} time steps $t$, exactly as simulations are commonly done following numerical integration schemes.
A set of observations or measurements $\mathbf{o}$ are available from sensors installed on the system.
Note, however, that the graphical model formalism is flexible enough
to allow each sensor to be available at a different or even sampling rate.
When using this graphical model to achieve state estimation, sensors
may provide \emph{partial} information on the system dynamics, which will be then fused with the rest of graph constraints over the system trajectory to reach the most-likely values of the estimated variables. In non state-estimation problems, the set of observations $\mathbf{o}$  may be empty.
Additionally, system parameters such as masses, stiffness or friction coefficients (constant or variable over time) could be added as additional nodes, although this possibility is left out of the scope of this manuscript for the sake of conciseness.

From \FIG{fig:dbn_org}, some observations can be drawn:
\begin{itemize}
 \item A central role in the DBN for independent coordinates (Figure~\ref{fig:dbn_org}(b)) is played by the independent coordinates ${\z,\dz,\ddz}$ as they represent the degrees of freedom (d.o.f) that govern the evolution of the MBS. This role is assumed by the dependent coordinate variables ${\q,\dq,\ddq}$ in Figure~\ref{fig:dbn_org}(a).

 \item The branch variable $C$, required uniquely in the independent-coordinate formulation, allows us to disambiguate between, e.g. the two possible configurations for a four-bar linkage if we are only given the information about the crank angle. This variable could be part of the unknowns to be estimated, as already done in \cite{blanco2015mds}, although that work did not use the more powerful FG representation proposed here.

 \item Typically, nodes in a DBN are depicted as shaded or unshaded depending of whether they are ``hidden'' or ``observed'' variables, respectively, e.g. see \cite{bishop2006pattern,blanco2015mds}.
 The former are estimated from the latter. Since this work focuses on FGs instead, where observed variables are subsumed into \emph{factor nodes} and unknowns become \emph{variable nodes} \cite{dellaert2017factor},  different FGs will be generated from the same DBNs in Figure~\ref{fig:dbn_org} for different multibody simulation problems, hence it is not convenient to establish such a distinction at the DBN level yet.

  \item Observations $\mathbf{o}$ (sensor readings) are a function of (all, or a subset of) dependent coordinates. Typical observations that may be useful in a MBD problem are measurements obtained from gyroscopes, accelerometers, and load cells.

 \item External forces $\F_t$ act by means of modifying accelerations $\ddq_t$ or $\ddz_t$, according to the system dynamics, expressed by \EQ{eq:ddq} or \EQ{eq:zDynSys}, respectively.

\end{itemize}

Once the model of the dynamics MBS has been expressed as a graphical model, the time evolution of the system can be obtained by converting the graph into a maximum-a-posteriori (MAP) estimation problem.
To this aim, one could write down the joint probability distribution $p(\phi)$ of all
the involved variables $\phi=\{\q_{0:t}, \dq_{0:t}, \ddq_{0:t}, \OBS_{0:t}, \F_{0:t}\}$ for time steps $0$ to $t$ exploiting the conditional probability encoded by the DBN (refer to e.g. \cite{bishop2006pattern,koller2009pgm}), which for dependent coordinates (i.e. \FIG{fig:dbn_org}(a)) becomes:

\begin{equation}
\label{eq:factorization.dbn}
p(\phi)
= p(\q_0) p(\dq_0)
\lp
 \prod_{i=0}^t p(\F_i)
\rp
\lp
 \prod_{i=1}^t
 p(\OBS_i|\q_i,\dq_i,\ddq_i)
 p(\ddq_i|\q_i,\dq_i,\F_i)
 p(\q_i|\q_{i-1},\dq_{i-1})
 p(\dq_i|\ddq_{i-1},\dq_{i-1})
\rp
\end{equation}

\noindent where each conditional probability term in \EQ{eq:factorization.dbn},
taking negative logarithms and up to an irrelevant proportionality constant, becomes a \emph{factor}
in this alternative form of an overall \emph{cost function} $c(\phi)$ to be minimized:

\begin{equation}
\label{eq:factorization.fg}
c(\phi)
= f_{prior}(\q_0)
f_{prior}(\dq_0)
\lp
 \prod_{i=0}^t f_{prior}(\F_i)
\rp
\lp
 \prod_{i=1}^t
 f_{obs}(\OBS_i,\q_i,\dq_i,\ddq_i)
 f^d_{dyn}(\ddq_i,\q_i,\dq_i,\F_i)
 f_{ei}(\q_i,\q_{i-1},\dq_{i-1})
 f_{ei}(\dq_i,\ddq_{i-1},\dq_{i-1})
\rp
\end{equation}

A fundamental feature of graphical models is how they enable
\emph{factorizing} functions of \emph{all} problem variables such as $c(\phi)\propto -\log p(\phi)$
into the product of a large number of \emph{smaller} functions
(in terms of the number of variables involved in each expression)
called \emph{factors} in \EQ{eq:factorization.fg},
which are discussed individually in section~\ref{sect:Factors}.
This is what keeps the estimation
problem tractable even for hundreds of thousands of variables, something intractable for
estimators in the family of the Kalman filter not exploiting the sparsity of the problem structure.
Note that the goal of an estimator is searching for the optimal set of unknowns $\phi^\star$
that maximize the likelihood of all observed data according to the model, that is:

\begin{equation}
\phi^\star = \arg\max_{\phi} p(\phi) = \arg\min_{\phi} c(\phi)
\end{equation}

Depending on the division of the DBN variables into observed and unknowns, we would arrive at different
factorizations since \emph{factors} are considered to be functions of the unknowns only.
For example, if all variables in one given cost factor in \EQ{eq:factorization.fg} are known data it
just evaluates to a constant which can be absorbed by the proportionality relationship
$c(\phi)\propto - \log p(\phi)$ and hence can be ignored.
A graphical representation of the remaining relevant factors leads us to \emph{factor graphs} themselves.

\section{Factor graphs for multibody dynamics}
\label{sect:fromMBDtoFGdep}

The conversion from DBN to FG is known to be achievable as follows,
without the need to explicitly writing down probabilistic equations
like \EQ{eq:factorization.dbn}--\EQ{eq:factorization.fg}:

\begin{quote}
Every Bayes net [DBN] node splits in both a variable node and a
factor node in the corresponding factor graph. The factor is connected
to the variable node, as well as the variable nodes corresponding to
the parent nodes in the Bayes net. If some nodes in the Bayes net are
evidence nodes, i.e., they are given as known variables, we omit the
corresponding variable nodes: the known variable simply becomes a
fixed parameter in the corresponding factor.
\hfill
\cite[p.~12]{dellaert2017factor}
\end{quote}

Applying these rules to the DBN reflecting the structure of variables involved in the dynamic simulation of multibody systems with dependent coordinates in \FIG{fig:dbn_org}(a) we derive next the FGs for
a couple of practical problems depending on which are the known and unknown variables.
Factors will be only briefly discussed here regarding their purpose and meaning, whereas their detailed implementations are described in section~\ref{sect:Factors}.

\subsection{FG for forward dynamics in dependent coordinates}
\label{sect:fg.forwardsDyn}

\begin{figure}
\centering
\includegraphics[width=0.65\linewidth]{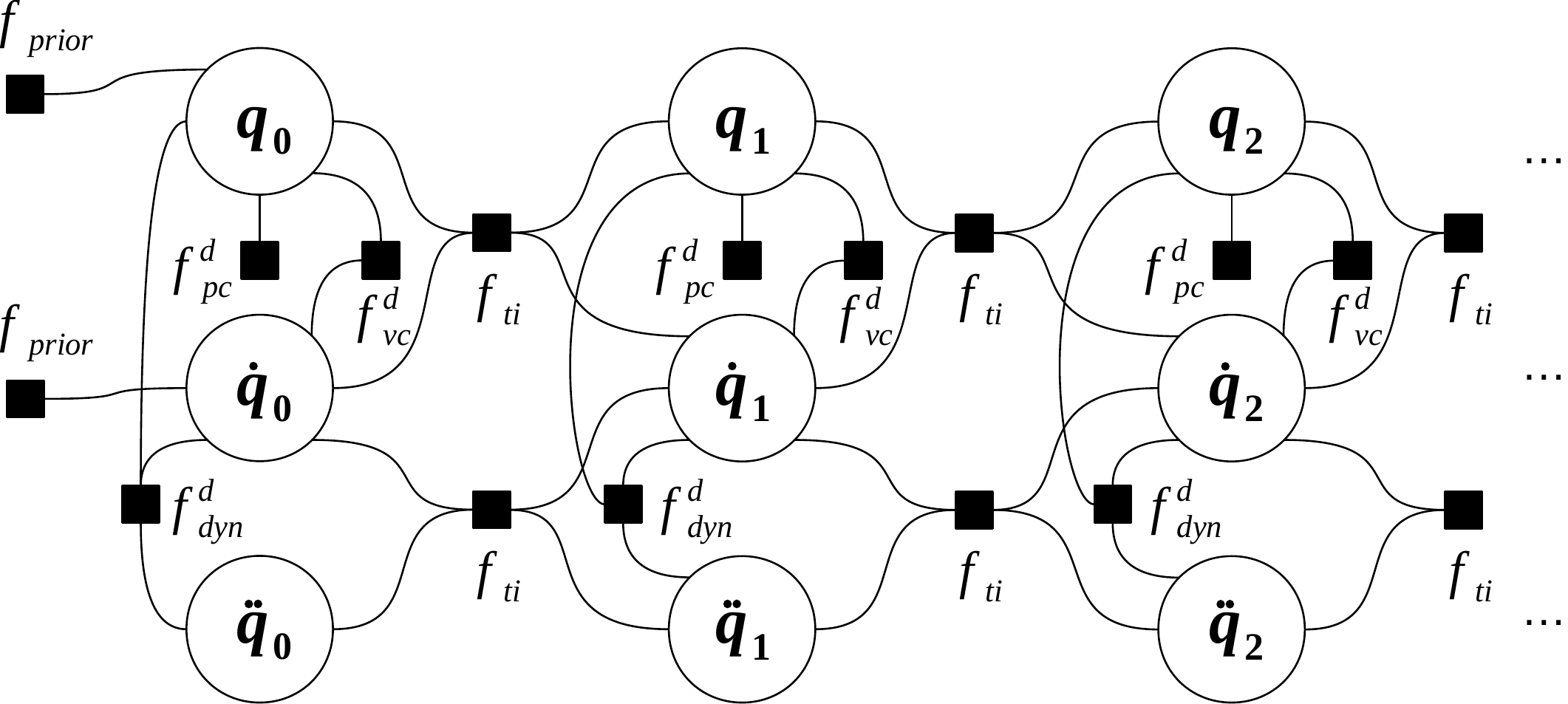}
\caption{Factor graphs for the forward dynamic simulation problem using dependent generalized coordinates. Circle nodes are problem unknowns, solid square nodes are factors. See discussion in section~\ref{sect:fg.forwardsDyn}.}
\label{fig:fg.dep.forwardDyn}
\end{figure}

In forward dynamics simulation, we are given a known initial state  of a mechanical system (position $\q_0$ and velocity $\q_0$), their geometric and inertial properties, and the external forces that act on it ($\F_i$, $i=0,...,N$). For simplicity, we assume no sensors are installed in the system since this problem instance does not need them, but predicting sensor outputs could be also possible adding the corresponding nodes and factors.

The resulting FG when using dependent coordinates is shown in \FIG{fig:fg.dep.forwardDyn} and it involves the following \emph{factor nodes} (or plain \emph{factors}):

\begin{itemize}
\item $f_{prior}$: Factors imposing a given \emph{a priori} knowledge about the attached variables, e.g. initial conditions. Prior factors can be defined for both, position and velocity.
% ----
\item $f^d_{pc}$: Factor for position constraints in dependent coordinates. It ensures the fulfillment of mechanical holonomic constraints by keeping the state vector $\q$ on the manifold $\PHI(\q,t)=\mathbf{0}$, hence this factor is repeated for each position node $\q_i$. It could be omitted for $\q_0$ if the enforced pose imposed by the prior factor is known to be a correct mechanism position complying with $\PHI(\cdot)=\mathbf{0}$; alternatively, the prior factor can be made to affect a minimum number of variables in $\q_0$ (the number of degrees of freedom), leaving $f^d_{pc}$ in charge of recalculating the rest of the generalized coordinates. This factor, on its own, solves the so-called \emph{position problem} \cite{garcia1994} in multibody dynamics.
% ----
\item $f^d_{vc}$: Factor for velocity constraints in dependent coordinates, enforcing generalized velocities $\dq$ to remain on the manifold $\Phiq \dq + \PHI_t=\mathbf{0}$.
% ----
\item $f^d_{dyn}$: Factor for the dynamics equation of motion: it links external forces (known data in this FG, hence not reflected as variable nodes) with the instantaneous acceleration $\ddq$. It also depends on $\q$ and $\dq$ since acceleration is always a function of them too. Note in the graph how acceleration for each timestep depends on position and velocity of the same timestep only.
% ----
\item $f_{ti}$: Trapezoidal numerical integrator factor is used twice per timestep: to integrate velocities into positions, and accelerations into velocities. Implicit integrators as the Trapezoidal one are often employed in MBS simulations for their stability. However, the Euler integrator version is also devised (see section~\ref{sect:Factors}), since explicit integrators are commonly used in real-time applications as well.
%We also devised an Euler integrator version (see section~\ref{sect:Factors}) but the trapezoidal version is more advisable since it ensures higher accuracy as a second-order integrator than the first-order Euler integration rule. \textcolor{red}{as long as actual accelerations are smooth..I do not like the previous definition in red}
\end{itemize}

\subsection{FG for forward dynamics in independent coordinates}
\label{sect:fg.forwardsDyn.indep}

\begin{figure}
\centering
\includegraphics[width=0.65\linewidth]{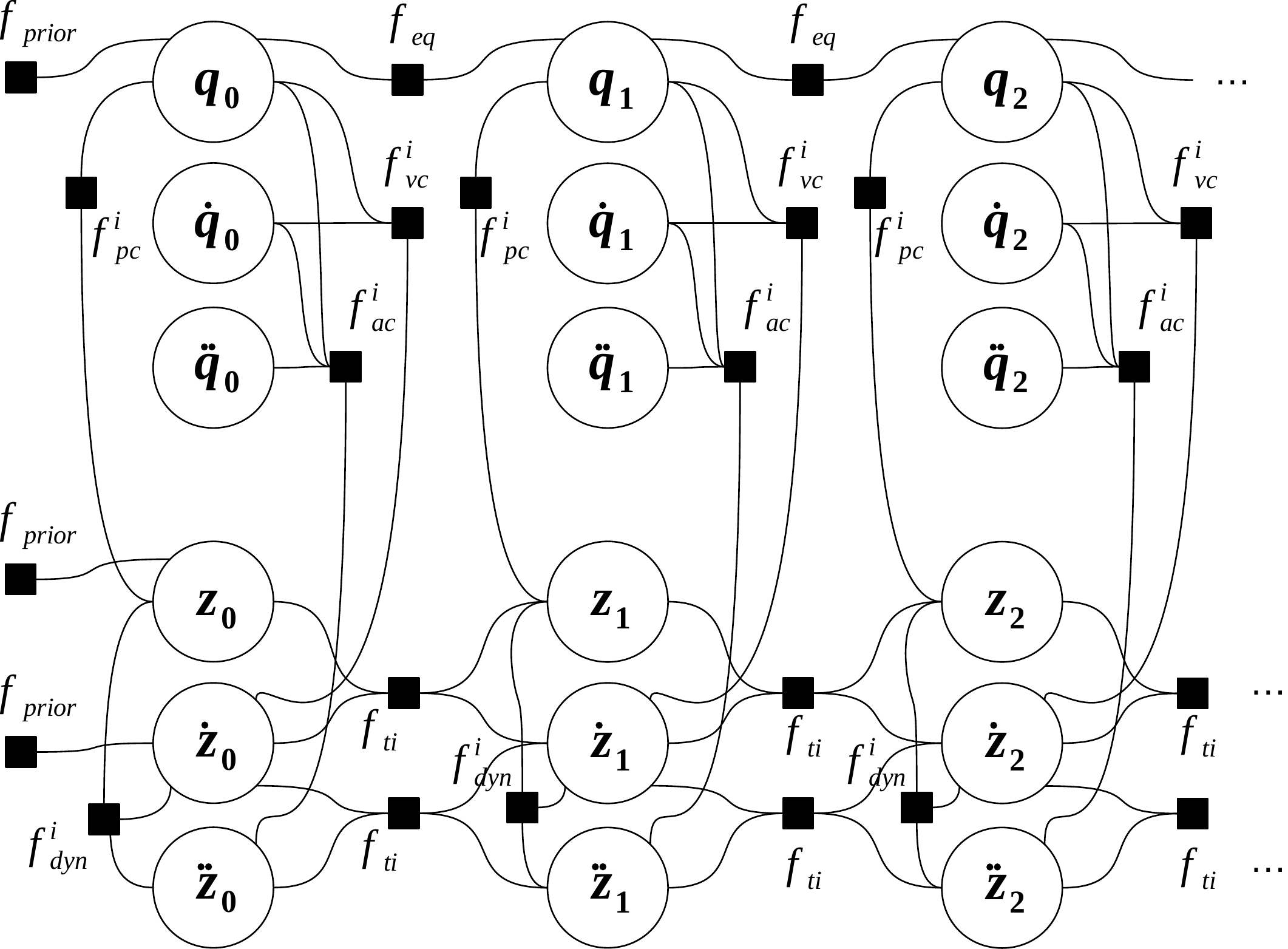}
\caption{Factor graph model for the forward dynamic simulation problem using independent generalized coordinates. Circle nodes are problem unknowns, solid square nodes are factors. See discussion in section~\ref{sect:fg.forwardsDyn.indep}.}
\label{fig:fg.indep.forwardDyn}
\end{figure}

Alternatively, one could devise a FG implementation for the forward dynamics problem when using \emph{independent coordinates},
leading to the graph depicted in \FIG{fig:fg.indep.forwardDyn}.
In this case, the following factors are used:
\begin{itemize}
\item $f_{prior}$: In this case, they are used to impose both, an initial known dynamic state ($\z_0$, $\dz_0$) and \emph{approximated} initial values for the full vector of dependent coordinates $\q_0$. The latter may be required to solve ambiguities in closed kinematic topologies, e.g. a four bar linkage, where knowing the minimum set of independent variables still leaves more than one feasible kinematic configuration.
% ----
\item $f_{eq}$: An equality factor, used to impose a \emph{soft} constraint between state vectors consecutive in time.
The rationale behind this factor is to impose a \emph{soft constraint}, which can be easily violated (its weight is small in comparison to all other factors) but still provides a solid starting point for nonlinear numerical solvers to exploit the fact that mechanism coordinates cannot vary abruptly between consecutive time steps. This factor is particularly important to solve ambiguities in mechanisms with more than one branch, exactly as argued by the end of the former paragraph.
% ----
\item $f^i_{dyn}$: The independent-coordinate version of $f^d_{dyn}$ discussed above, it implements the dynamics equation of motion.
\item $f^i_{pc}$, $f^i_{vc}$: Like their dependent-coordinate counterparts in the former section, these factors keep the position and velocity vector within their corresponding constraint manifolds. Note how, in this case, the factors not only affect $\q$ and $\dq$ coordinates, but their independent-coordinate versions $\z$ and $\dz$ too, respectively. These factors actually correspond to the so-called \emph{position problem} and \emph{velocity problem} in multibody mechanics \cite{garcia1994}.
% ---
\end{itemize}

\subsection{FG for inverse dynamics in dependent coordinates}
\label{sect:fg.inverseDyn}

\begin{figure}
\centering
\includegraphics[width=0.65\linewidth]{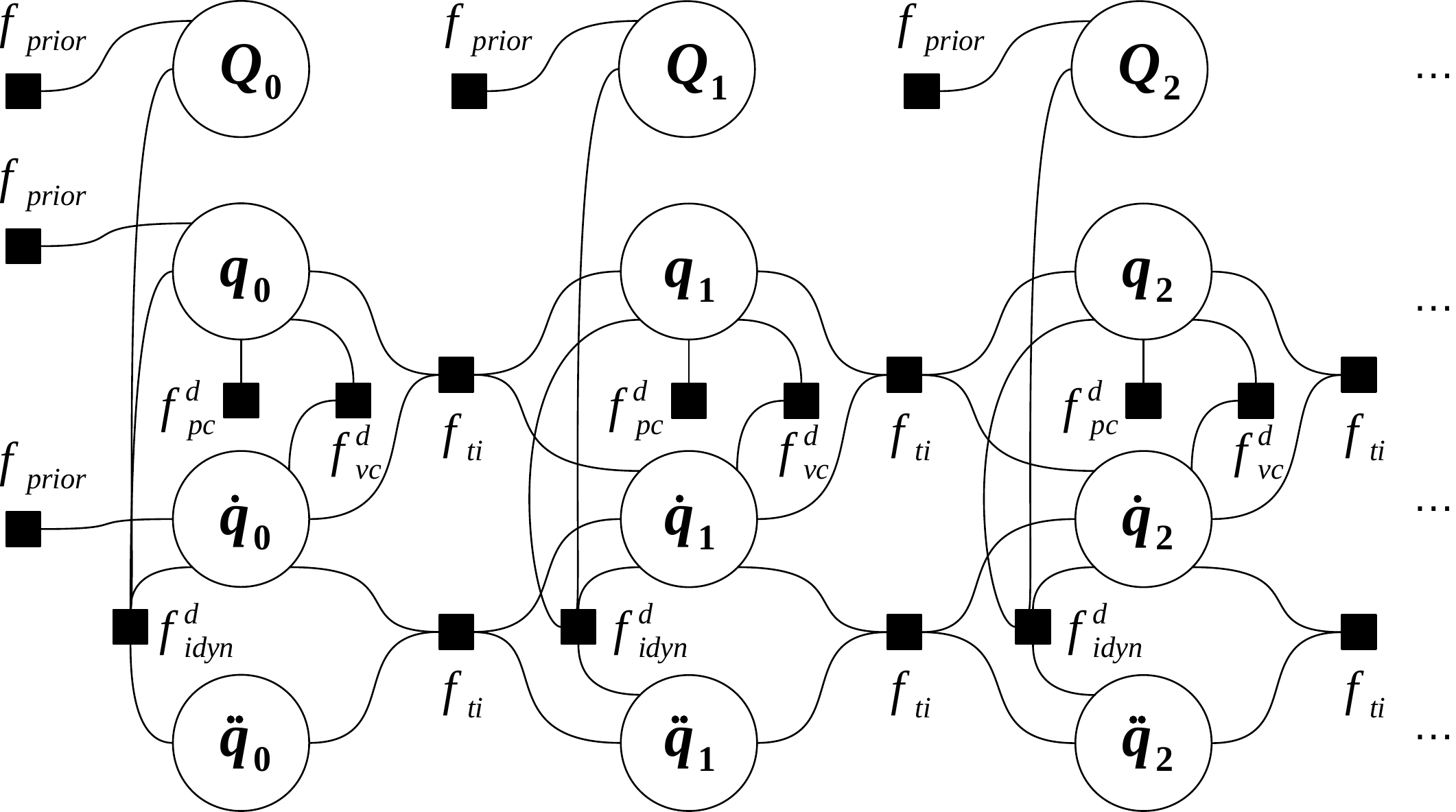}
\caption{Factor graph for the inverse dynamics problem using dependent generalized coordinates. See discussion in section~\ref{sect:fg.inverseDyn}.}
\label{fig:fg.inverseDyn}
\end{figure}

Another problem that can be formalized as a FG is inverse dynamics, as shown in \FIG{fig:fg.inverseDyn}.
The problem consists of specifying a desired trajectory over time for the set of degrees of freedom of a mechanism,
then solving for the required forces and torques to generate such trajectory.
The following factors are used:
\begin{itemize}
\item $f_{prior}$: Here, different prior factors are used to define initial known values for the dynamic state ($\q_0$, $\dq_0$)
and to enforce a value of zero in all components of the generalized force vectors $\F_i$ where it is known that no external force is acting. In other words, the latter factors are required to leave only part of $\F_i$ as an unknown, e.g. for the degree of freedom where a motor is exerting a torque.
% ----
\item $f^d_{idyn}$: Similar to the dynamics equation factor $f^d_{dyn}$ discussed above, but with the force $\F_i$ as an additional variable. Note that $f^d_{dyn}$ (see section~\ref{sect:fg.forwardsDyn.indep}) also used a \emph{value} for $\F_i$ but it was treated as a known constant, whereas for $f^d_{idyn}$ the $\F_i$ are unknowns and as such the factor provides a Jacobian of the error term with respect to them as well.
\end{itemize}

From all discussed problems so far,
inverse dynamics is the hardest to solve
for nonlinear iterative solvers from
initial values that are far from the optimum.
Therefore, it requires a proper initialization strategy to enable numerical nonlinear solvers to cope with it effectively, as discussed in the experiments section later on.

%We propose solving it using a 4-stage procedure (Explain in experiments...).

% ==================== factors ====================
\section{Factors definition}
\label{sect:Factors}

In a factor graph, factors establish the relations between variables.  This section provides an insightful practical guide to those factors MBS problems are made of.

Each factor defines an \emph{error} $\e(\x)$ between predicted and measured data. To apply nonlinear optimization algorithms (e.g. Gauss-Newton or Levenberg-Marquardt), the Jacobian of such error with respect to all involved variables $\x$ is required.
DBN variables whose values are known do not become FG nodes
and hence are \emph{constant parameters} of factors.
The goal of the optimization is to search for the best values of all variables $\x^\star$ taking the weighted sum of error functions (one per factor) as close to zero as possible:

\begin{equation}
\x^\star = \arg \min_{\x} \sum_{i} \norm{\e_i(\x)}^2_{\boldsymbol{\Sigma}}
\label{eq:nonlinear-LS}
\end{equation}

\noindent where $\norm{\e_i(\x)}^2_{\boldsymbol{\Sigma}}$ is a form of \emph{whitening} already integrating the probabilistic \emph{noise model} (or \emph{weight})
of each factor.
The most common model, used in the present work, is the assumption
of additive zero mean Gaussian noise $\mathbf{n} \sim \mathcal{N}(\textbf{0}, \boldsymbol{\Sigma})$
with covariance matrix $\boldsymbol{\Sigma}$.
Taking the negative logarithm of the corresponding probability density can be shown to give us the \emph{nonlinear least-squares} problem in Eq.(\ref{eq:nonlinear-LS}), where

\begin{equation}
\norm{\e(\x)}^2_{\boldsymbol{\Sigma}} = \frac{1}{2} \e(\x)^\top \boldsymbol{\Lambda} \e(\x)
\end{equation}

\noindent with information matrix $\boldsymbol{\Lambda}=\boldsymbol{\Sigma}^{-1}$.
Larger information values in $\boldsymbol{\Lambda}$ (or smaller variances in $\boldsymbol{\Sigma}$) imply that the factor must be considered with a higher weight during the optimization
in comparison to other factors with smaller information values (or larger variances).
Note that each component of multidimensional $\x$ vectors may have a different weight for a given single factor, e.g. as proposed in the priors over external forces in section~\ref{sect:fg.inverseDyn}.
The interested reader is deferred to \cite{dellaert2017factor} for a more in-depth discussion on this topic.

In the following,
note the use of the superscripts $f^{d}_{\cdot}$ and $f^{i}_{\cdot}$ for factors
with differentiated implementations for dependent and independent-coordinates, respectively.

\subsection{$f_{prior}$: Prior factor}
\label{sect:factor.prior}

Prior factors $f_{prior}(\x)$ over a variable $\x$ are the only unary factor used in our proposed graphs.
They represent \emph{a priori} belief (hence the name) about the state of a given variable.

Since the problems addressed in this manuscript use variables that are either (a) state vectors with generalized coordinates of planar mechanisms, or (b) their velocities or (c) their accelerations, all variables can be treated as vectors in the group of real numbers, hence the error function $\e(\cdot)$ of this factor for a vector of dimensionality $d$ has a simple form:

\begin{subeqnarray}
\label{eq:factor.prior}
\text{$f_{prior}$ error function:} &&
\e(\x) = \x - \x_0 \\
\text{Variables:} && \x \\
\text{Fixed parameters:} && \x_0  \\
\text{Jacobian:} && \dfrac{\partial \e}{\partial \x} = \mathbf{I}_d
\end{subeqnarray}

\noindent where $\x_0$ is the ``desired'' value for the variable $\x$.

\begin{figure*}
\centering
	\begin{subfigure}[t]{0.17\textwidth}
		\includegraphics[width=1\textwidth]{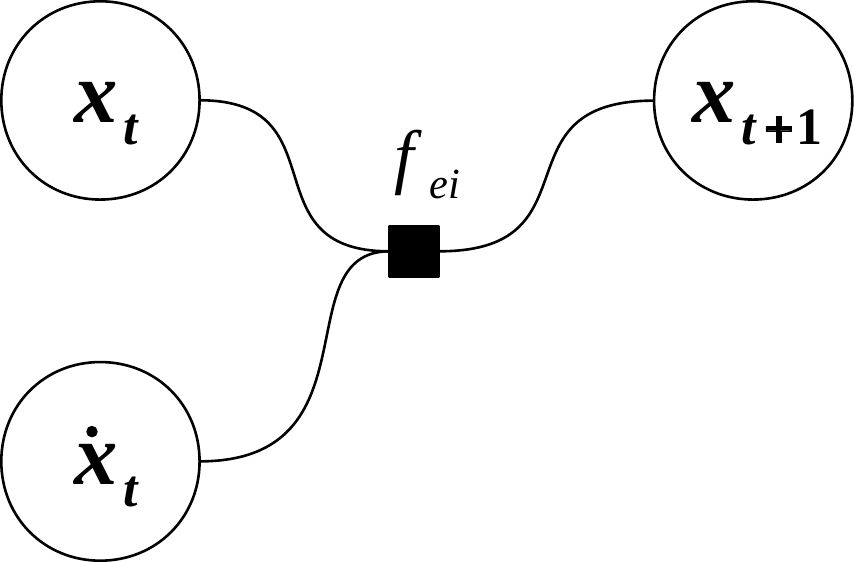}
		\caption{Euler integration.}
	\end{subfigure}
	\hspace*{1cm}
	\begin{subfigure}[t]{0.17\textwidth}
		\includegraphics[width=1\textwidth]{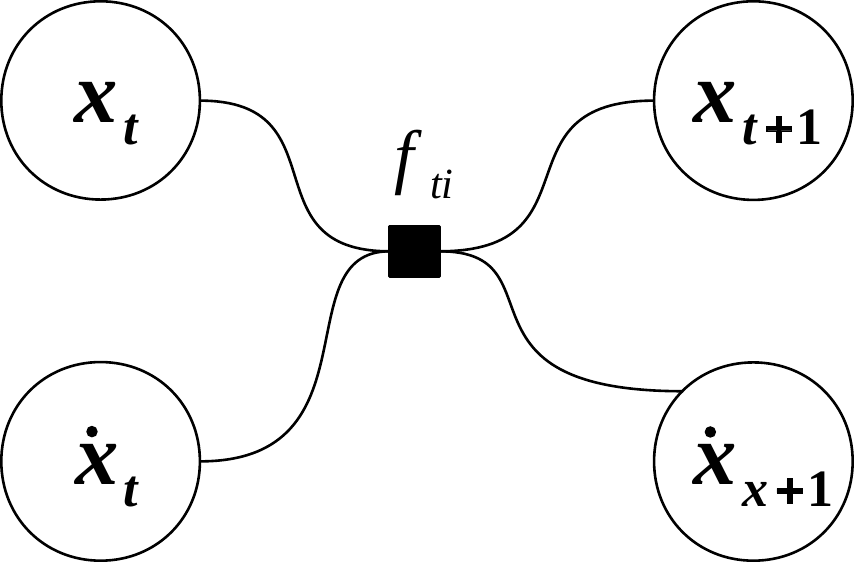}
		\caption{Trapezoidal integration.}
	\end{subfigure}
	\hspace*{1cm}
	\begin{subfigure}[t]{0.10\textwidth}
		\centering
		\includegraphics[width=0.5\textwidth]{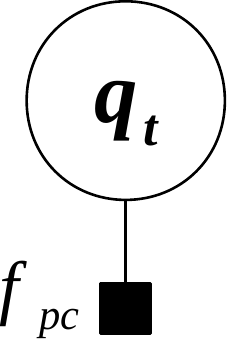}
		\caption{Position constraints}
	\end{subfigure}
	\hspace*{1cm}
	\begin{subfigure}[t]{0.16\textwidth}
		\centering
		\includegraphics[width=0.5\textwidth]{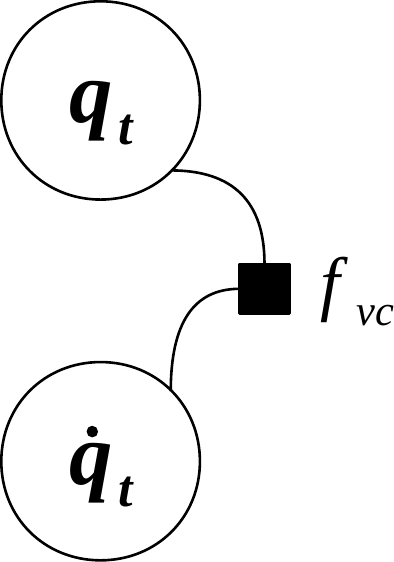}
		\caption{Velocity constraints}
	\end{subfigure}	
\caption{(a)--(b) Numerical integration factors discussed in sections~\ref{sect:factor.euler.int}-\ref{sect:factor.trap.int}.
Factors used to enforce $\q$ and $\dq$ to remain within their corresponding manifolds: (c) holonomic position constraints, (b) velocity constraints (see sections~\ref{sect:factor.pos.constr.dep}-\ref{sect:factor.vel.constr.dep}).
}
\label{fig:individual.factors}
\end{figure*}

\subsection{$f_{ei}$: Euler integrator factor}
\label{sect:factor.euler.int}

The graphs proposed in former sections work over \emph{temporal sequences} of variables, sampled at a fixed sample rate $f_s=1/\Delta t$.
Imposing the continuity of ordinary differential equations in discrete time
can be done by means of numerical integration, of which the Euler integrator is
the simplest instance.

Given two consecutive samples for time steps $t$ and $t+1$ of a given variable $\x$ and its time derivative $\dx$ at time $t$, the Euler integrator factor $f_{ei}$ (shown in \FIG{fig:individual.factors}(a)) can be defined as:

\begin{subeqnarray}
\label{eq:factor.fei}
\text{$f_{ei}$ error function:} &&
\e(\x_t,\x_{t+1},\dx_t) = \x_{t+1} - \x_t - \Delta t \dx_t \\
\text{Variables:} && \x_t,\x_{t+1},\dx_t \\
\text{Fixed parameters:} && \Delta t  \\
\text{Jacobians:}
&& \dfrac{\partial \e}{\partial \x_t} = -\mathbf{I}_d
\quad
 \dfrac{\partial \e}{\partial\x_{t+1}} =  \mathbf{I}_d
\quad
 \dfrac{\partial \e}{\partial \dx_t} =  -\Delta t \mathbf{I}_d
\end{subeqnarray}

\noindent where the state space is assumed to be $\mathbb{R}^d$.

\subsection{$f_{ti}$: Trapezoidal integrator factor}
\label{sect:factor.trap.int}

Another well-known numerical integration scheme follows the Trapezoidal Rule. Given two consecutive samples for time steps $t$ and $t+1$, both for a given variable $\x$ and its time derivative $\dx$, the Trapezoidal integrator factor $f_{ti}$, shown in \FIG{fig:individual.factors}(b), can be defined as:

\begin{subeqnarray}
\label{eq:trapfactor.}
\text{$f_{ti}$ error function:} &&
\e(\x_t,\x_{t+1},\dx_t,\dx_{t+1}) = \x_{t+1} - \x_t - \dfrac{\Delta t}{2} \dx_t  - \dfrac{\Delta t}{2} \dx_{t+1} \\
\text{Variables:} && \x_t,\x_{t+1},\dx_t,\dx_{t+1} \\
\text{Fixed parameters:} && \Delta t  \\
\text{Jacobians:}
&& \dfrac{\partial \e}{\partial \x_t} = -\mathbf{I}_d
\quad
 \dfrac{\partial \e}{\partial\x_{t+1}} =  \mathbf{I}_d
\quad
 \dfrac{\partial \e}{\partial \dx_t} =  -\dfrac{\Delta t}{2} \mathbf{I}_d
\quad
 \dfrac{\partial \e}{\partial \dx_{t+1}} =  -\dfrac{\Delta t}{2} \mathbf{I}_d
\end{subeqnarray}

As mentioned in section~\ref{sect:fg.forwardsDyn}, this integrator is preferred for its better accuracy in comparison to the Euler method.
%Note that the same integration factor plays a dual role, integrating positions or velocities depending on whether $\lp \x,\dx\rp $ are $\lp \q,\dq\rp $ or $\lp \dq,\ddq\rp$ respectively, for the dependent coordinates (similar for the independent ones).

\subsection{$f^d_{pc}$: Factor for position constraints in dependent coordinates}
\label{sect:factor.pos.constr.dep}

This factor, depicted in \FIG{fig:individual.factors}(c), ensures the fulfillment of mechanical holonomic constraints of \EQ{eq:Constraints}.
As explained in Section~\ref{sect:DepCoordFormulation}, modeling a mechanism
in dependent coordinates leads to a number of constraint equations
largest or equal to the number of generalized coordinates.

This factor has the following general form:

\begin{subeqnarray}
\label{eq:factor_fd_pc}
\text{$f^d_{pc}$ error function:} &&
\e(\q_t) = \PHI(\q_t)  \\
\text{Variables:} && \q_t \\
\text{Fixed parameters:} && \text{(constant distances, angles, etc.)} \\
\text{Jacobians:} && \dfrac{\partial \e}{\partial \q_t} = \Phiq \lp \q_t \rp
\end{subeqnarray}

\noindent where both $\PHI$ and $ \Phiq$ can be automatically built out of pre-designed blocks (see Appendix~III), according to a formal description
of the topology of the mechanical system and the specific joints connecting each pair of adjacent bodies.
In particular, since each physical constraint
becomes one or more entries in the $\PHI(\q)$ vector,
each such constraint fully determines the corresponding rows
in the Jacobian $\Phiq(\q)$, easing its automated construction
from the elemental expressions exposed in Appendix~III.

\subsection{$f^d_{vc}$: Factor for velocity constraints in dependent coordinates}
\label{sect:factor.vel.constr.dep}

Velocity constraint factors further improve the quality of MBS simulation. These kind of factors are modeled by \EQ{eq:ConstraintsVel} and depend on both, $\q$ and $\dq$, as illustrated \FIG{fig:individual.factors}(d). Its definition is as follows:

\begin{subeqnarray}
	\label{eq:factor_fd_vc}
	\text{$f^d_{vc}$ error function:} &&
	\e(\q_t, \dq_t) = \Phiq(\q_t) \dq_t - \cancelto{0}{\Phi_t(\q_t)} = \Phiq(\q_t) \dq_t \\
	\text{Variables:} && \q_t, \dq_t \\
	\text{Fixed parameters:} && \text{(constant distances, angles, etc.)} \\
	\text{Jacobians:} &&
	\dfrac{\partial \e}{\partial \q_t} =
	\dfrac{\partial \Phiq \lp \q_t \rp \dq_t}{\partial \q_t}
	\quad
	\dfrac{\partial \e}{\partial \dq_t} =
	\Phiq(\q_t)	
\end{subeqnarray}

\noindent where we assumed that no constraint depends explicitly on time, hence $\PHI_t(\q_t)$ can be safely neglected.
Again, each row in the Jacobians comes from exactly one constraint in the
definition of the mechanical system, with most common cases
described in Appendix~III.

\subsection{$f^d_{dyn}$, $f^d_{idyn}$ and $f^i_{dyn}$, $f^i_{idyn}$: Factors for dynamics}

These factors minimize the error between the actual acceleration estimates ($\HAT{\ddq}$ and $\ddz$) and the corresponding equations of motion \EQ{eq:ddq} and \EQ{eq:zDynSys} for dependent and independent coordinates, respectively.
To deal both forward and inverse dynamics, this factor connects the generalized positions, velocities, and accelerations with the forces ($\Q_t$) for each single time step $t$.
In the case of forward dynamics, both the variable $\Q_t$ and the edge connecting it to the factor are dropped, becoming $\Q_t$ an internal known parameter to the factor. For this reason, two factors for forward and inverse dynamics must be defined independently:

\begin{itemize}
\item \textbf{Forward dynamics}

For dependent coordinates:

\begin{subeqnarray}
\label{eq:FactorDepDyn.dir}
\text{$f^d_{dyn}$ error function:} &&
\e(\q_t,\dq_t,\ddq_t) = \ddq(\q_t,\dq_t)-\ddq_t \\
\text{Variables:} &&
\q_t, \dq_t, \ddq_t \\
\text{Fixed parameters:} && (\text{masses, external forces})  \\
\text{Jacobians:} &&
\dfrac{\partial\e}{\partial\q_t} = \dfrac{\partial\ddq(\q_t,\dq_t)}{\partial\q_t} \quad \dfrac{\partial\e}{\partial\dq_t} = \dfrac{\partial\ddq(\q_t,\dq_t)}{\partial\dq_t} \quad \quad \dfrac{\partial\e}{\partial\ddq_t} = -\mathbf{I}
\end{subeqnarray}

\noindent and for independent coordinates:

\begin{subeqnarray}
\label{eq:FactorIndepDyn.dir}
\text{$f^i_{dyn}$ error function:} &&
\e(\z_t,\dz_t,\ddz_t) = \ddz(\z_t,\dz_t)-\ddz_t \\
\text{Variables:} &&
\z_t, \dz_t, \ddz_t \\
\text{Fixed parameters:} && (\text{masses, external forces})  \\
\text{Jacobians:} &&
\dfrac{\partial\e}{\partial\z_t} = \dfrac{\partial\ddz(\z_t,\dz_t)}{\partial\z_t} \quad \dfrac{\partial\e}{\partial\dz_t} = \dfrac{\partial\ddz(\z_t,\dz_t)}{\partial\dz_t} \quad \quad \dfrac{\partial\e}{\partial\ddz_t} = -\mathbf{I}
\end{subeqnarray}

\item \textbf{Inverse dynamics}

For dependent coordinates:

\begin{subeqnarray}
\label{eq:FactorDepDyn.inv}
\text{$f^d_{dyn}$ error function:} &&
\e(\q_t,\dq_t,\ddq_t,\Q_t) = \ddq(\q_t,\dq_t,\Q_t)-\ddq_t \\
\text{Variables:} &&
\q_t, \dq_t, \ddq_t, \Q_t \\
\text{Fixed parameters:} && (\text{masses})  \\
\text{Jacobians:} &&
\dfrac{\partial\e}{\partial\q_t} = \dfrac{\partial\ddq(\q_t,\dq_t,\Q_t)}{\partial\q_t} \quad \dfrac{\partial\e}{\partial\dq_t} = \dfrac{\partial\ddq(\q_t,\dq_t,\Q_t)}{\partial\dq_t} \nonumber \\ &&
\dfrac{\partial\e}{\partial\ddq_t} = -\mathbf{I} \quad \dfrac{\partial\e}{\partial\Q_t} = \dfrac{\partial\ddq(\q_t,\dq_t,\Q_t)}{\partial\Q_t}
\slabel{eq:FactorDepDyn.inv.jacob}
\end{subeqnarray}

\noindent and for independent coordinates:

\begin{subeqnarray}
\label{eq:FactorIndepDyn.inv}
\text{$f^i_{dyn}$ error function:} &&
\e(\z_t,\dz_t,\ddz_t,\Q_t) = \ddz(\z_t,\dz_t,\Q_t)-\ddz_t \\
\text{Variables:} &&
\z_t, \dz_t, \ddz_t, \Q_t \\
\text{Fixed parameters:} && (\text{masses})  \\
\text{Jacobians:} &&
\dfrac{\partial\e}{\partial\z_t} = \dfrac{\partial\ddz(\z_t,\dz_t,\Q_t)}{\partial\z_t} \quad \dfrac{\partial\e}{\partial\dz_t} = \dfrac{\partial\ddz(\z_t,\dz_t,\Q_t)}{\partial\dz_t} \nonumber \\ &&
\dfrac{\partial\e}{\partial\ddz_t} = -\mathbf{I} \quad \dfrac{\partial\e}{\partial\Q_t} = \dfrac{\partial\ddz(\z_t,\dz_t,\Q_t)}{\partial\Q_t}
\slabel{eq:FactorIndepDyn.inv.jacob}
\end{subeqnarray}

\end{itemize}

%\begin{equation}
%\mathbf{r}^i_{em}(\HAT{\ddq}_i,\dq_i,\q_i, \Q_i) =  \ddq(\q_i,\dq_i, \Q_i)-\HAT{\ddq}_i
%\label{eq:ddq_err}
%\end{equation}
%where $\mathbf{r}^i_{em}$ is the \textit{dynamic error} of the \textit{equation of motion} at the istant $i$. Therefore, its Jacobian is
%
%\begin{equation}
%\dfrac{\partial \mathbf{r}^i_{em}(\HAT{\ddq}_i,\dq_i,\q_i, \Q_i)}{\partial (\HAT{\ddq}_i,\dq_i,\q_i, \Q_i)} =
%\left[
%\dfrac{\partial \ddq(\q_i,\dq_i, \Q_i)}{\partial \q_i} \hspace*{0.25cm}
%\dfrac{\partial \ddq(\q_i,\dq_i, \Q_i)}{\partial \dq_i} \hspace*{0.25cm}
%-\mathbf{I}
%\hspace*{0.25cm}
%\dfrac{\partial \ddq(\q_i,\dq_i, \Q_i)}{\partial \Q_i} \right]
%\label{eq:qDynJacobian}
%\end{equation}

%Refer to \ref{sect:Apx.Jacobs.Dep.Coord} and \ref{sect:Apx.Jacobs.Indep.Coord} for details on the closed-form Jacobians in \EQ{eq:FactorDepDyn.dir}d and \EQ{eq:FactorDepDyn.inv}d,
%and to

Refer to \EQ{eq:ddq} and \EQ{eq:zDynSys} for the evaluation of the error functions above.
Numerical Jacobians have been used in our implementation for \EQ{eq:FactorDepDyn.inv.jacob} and \EQ{eq:FactorIndepDyn.inv.jacob},
since exact closed-form Jacobians are so complex than
the computational advantage of having closed-form expressions. In this particular case is not clear and deserves future analysis.

\subsection{$f^i_{pc}$: Factor for position constraints in independent coordinates}
\label{sect:factor.pos.constr.indep}

\FIG{fig:fg.indep.forwardDyn} shows three sets of factors
linking dependent coordinates (position, velocities, and accelerations) with their corresponding degrees of freedom,
that is, with their counterparts in independent coordinates.
The first one of these factor is $f^i_{pc}$,
in charge of imposing the simultaneous fulfillment of:
(a) position constraints in $\PHI(\q)$,
and (b) that independent coordinates $\z$ are integrated
into its corresponding positions within $\q$.
Therefore, its error function and Jacobians read:

\begin{subeqnarray}
\label{eq:factor_fi_pc}
\text{$f^i_{pc}$ error function:} &&
\e(\q_t, \z_t) =
 \begin{bmatrix}
  \PHI(\q_t) \\
%  \hline
  \q_t(\{idxs\}) - \z_t
 \end{bmatrix}_{(m+d) \times 1} \\
\text{Variables:} && \q_t, \z_t \\
\text{Fixed parameters:} && \{idxs\} \text{, constant distances, angles, etc.} \\
\text{Jacobians:} && \dfrac{\partial \e}{\partial \q_t} =
 \begin{bmatrix}
  \Phiq \lp \q_t \rp \\
%  \hline
  \mathbf{I}_{idxs}
 \end{bmatrix}_{(m+d) \times n}
 \quad
 \dfrac{\partial \e}{\partial \z_t} =
  \begin{bmatrix}
     \mathbf{0}_{m \times d} \\
%   \hline
     -\mathbf{I}_{d} \\
  \end{bmatrix}_{(m+d) \times d}
\end{subeqnarray}

\noindent where $\{idxs\}=\{y_1,y_2,...,y_d\}$ stands for the fixed sequence of indices of each $\z$ coordinate inside the $n$-vector $\q$, and the coefficients $I_{i,j}$ of the binary matrix $\mathbf{I}_{idxs}$ are defined as $1$ if $j=y_i$, or as $0$ otherwise,
where $i={1,...,d}$ and $j={1,...,n}$.

\subsection{$f^i_{vc}$: Factor for velocity constraints in independent coordinates}
\label{sect:factor.vel.constr.indep}

This factor ensures the fulfillment of the velocity constraints for independent coordinates:

\begin{subeqnarray}
	\label{eq:factor_fi_vc}
	\text{$f^i_{vc}$ error function:} &&
	\e(\q_t, \dq_t, \dz_t) =
	\begin{bmatrix}
  \Phiq(\q_t)\dq_t \\
%  \hline
  \dq_t(\{idxs\}) - \dz_t
 \end{bmatrix}_{(m+d) \times 1} \\
	\text{Variables:} && \q_t, \dq_t, \dz_t \\
	\text{Fixed parameters:} && \{idxs\} \text{, constant distances, angles, etc.} \\
	\text{Jacobians:} &&
	\dfrac{\partial \e}{\partial \q_t} =
	\begin{bmatrix}
		\Phiqq(\q_t) \dq_t \\
		\mathbf{0}_{d \times n}
	\end{bmatrix}_{(m+d) \times n}
	\quad
	\dfrac{\partial \e}{\partial \dq_t} =
	\begin{bmatrix}
		\Phiq(\q_t) \\
		\mathbf{I}_{idx}
	\end{bmatrix}_{(m+d) \times n}
	\quad
	\dfrac{\partial \e}{\partial \dz_t} =
  \begin{bmatrix}
     \mathbf{0}_{m \times d} \\
%   \hline
     -\mathbf{I}_{d} \\
  \end{bmatrix}_{(m+d) \times d}
\end{subeqnarray}

\noindent where the tensor-vector product $\Phiqq \dq_t$ results
in the 2-D matrix $\dot{\PHI}_\q$ (see \EQ{eq:apx.common.mbs.a}) which can be systematically built row by row from a description
of the mechanism from the building blocks described in Appendix~III.

\subsection{$f^i_{ac}$: Factor for acceleration constraints in independent coordinates}
\label{sect:factor.acc.constr.indep}

This factor ensures the fulfillment of the acceleration constraints for independent coordinates:

\begin{subeqnarray}
	\label{eq:factor_fi_ac}
	\text{$f^i_{vc}$ error function:} &&
	\e(\q_t, \dq_t,\ddq_t, \ddz_t) =
	\begin{bmatrix}
		\dot{\PHI}_{\q}(\q_t)\dq_t + \Phiq(\q_t)\ddq_t \\
		\ddq_t(\{idxs\}) - \ddz_t
	\end{bmatrix}_{(m+d) \times 1}	 \\
	\text{Variables:} && \q_t, \dq_t, \ddq_t, \ddz_t \\
	\text{Fixed parameters:} && \text{(constant distances, angles, etc.)} \\
	\text{Jacobians:} &&
	\dfrac{\partial \e}{\partial \q_t} =
	\begin{bmatrix}
		\dotPhiqq(\q_t) \dq_t + \Phiqq(\q_t)
		       \ddq_t \\
		\mathbf{0}_{d \times n}
	\end{bmatrix}_{(m+d) \times n}
	\quad
	\dfrac{\partial \e}{\partial \dq_t} =
	\begin{bmatrix}
		2\Phiqq(\q_t)\dq_t \\
		\mathbf{0}_{d \times n}
	\end{bmatrix}_{(m+d) \times n}	\nonumber \\ &&
	\dfrac{\partial \e}{\partial \ddq_t} =	
	\begin{bmatrix}
		\Phiq(\q_t) \\
		\mathbf{I}_{idx}
	\end{bmatrix}_{(m+d) \times n}
	\quad	
	\dfrac{\partial \e}{\partial \ddz_t} =
    \begin{bmatrix}
      \mathbf{0}_{m \times d} \\
%   \hline
     -\mathbf{I}_{d} \\
    \end{bmatrix}_{(m+d) \times d}
\end{subeqnarray}

\noindent where the tensor-vector products $\dotPhiqq(\q_t) \dq_t$, $\Phiqq \ddq_t$, and $2\Phiqq \dq_t = 2\dot{\PHI}_\q$ can be systematically built as described in Appendix~III.

% ==================== Applications ====================
\section{Test case validation}
\label{sect:Experiments}
\begin{figure}[t]
	\centering
	\begin{subfigure}[t]{0.3\textwidth}
	\centering
	\includegraphics[width=1\textwidth]{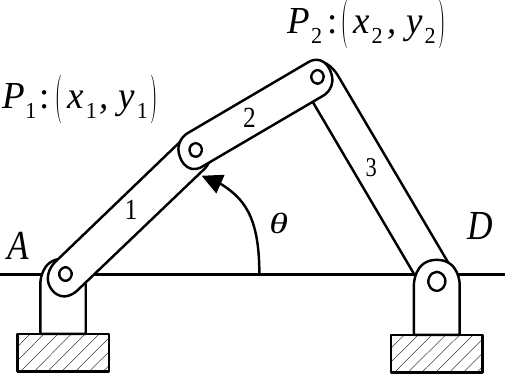}
	\caption{}
	\end{subfigure}	
	% --
	\begin{subfigure}[t]{0.40\textwidth}
	\centering
\vspace{-4cm}
\begin{tabular}{|c|rl|}
\hline
$L_1$ & 1.0 & [m] \\
$L_2$ & 2.0 & [m] \\
$L_3$ & $\sqrt{13}$ & [m] \\
$m_1$ & 1.0 & [kg] \\
$m_2$ & 2.0 & [kg] \\
$m_3$ & 4.0 & [kg] \\
$I_{i,g}$ & $\frac{1}{12}m_i L_i^2$ & [kg $\text{m}^2$] \\
\hline
\end{tabular}
	\caption{}
	\end{subfigure}
	% --
	\caption{(a) The planar mechanism used in the numeric simulations described in section~\ref{sect:Experiments}.
	(b) Table of mechanism properties (length and mass) for each link. Inertias $I_i$ are given with respect to the center of mass of each link, placed at its center.}
	\label{fig:4bar}
\end{figure}

A four-bar linkage is employed to exemplify the implementation and the performance of the proposed FG-based framework.
A scheme of the mechanism is shown in \FIG{fig:4bar}(a), where the two revolute joints, $P_1$ and $P_2$, can move in the motion plane, whereas $A$ and $D$ are fixed.
Geometric and inertial properties are summarized in \FIG{fig:4bar}(b).
The motion of the mechanism has been simulated using the commercial MBS environment Adams/View from \textit{MSC.Software} with a fixed-step integrator at $1/10$ of the time step used in the factor graph, hence its results can be considered the ground-truth for comparison purposes.

A companion open-source reference C++ implementation\footnote{In \url{https://github.com/MBDS/multibody-state-estimation}}
of the proposed factor graph approach has been released along with the paper.
Note that all factor closed-form Jacobians have been thoroughly validated by means of unit tests against numerical finite differences.

\begin{figure*}
	\centering
	\begin{subfigure}[t]{0.46\textwidth}
		\includegraphics[width=1\textwidth]{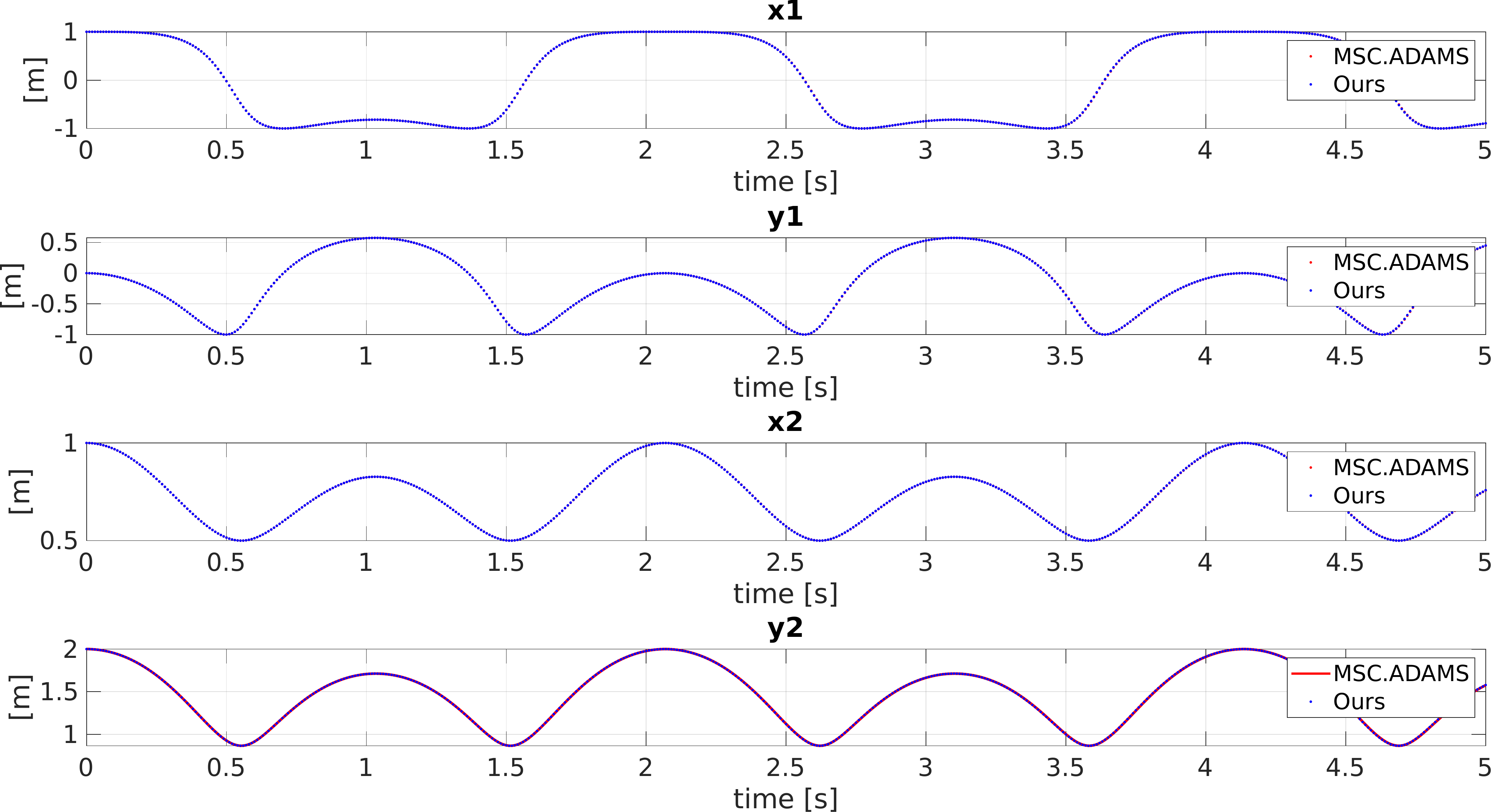}
		\caption{Comparison of $\q$ series.}
	\end{subfigure}
	\hspace*{0.5cm}
	\begin{subfigure}[t]{0.46\textwidth}
	\includegraphics[width=1\textwidth]{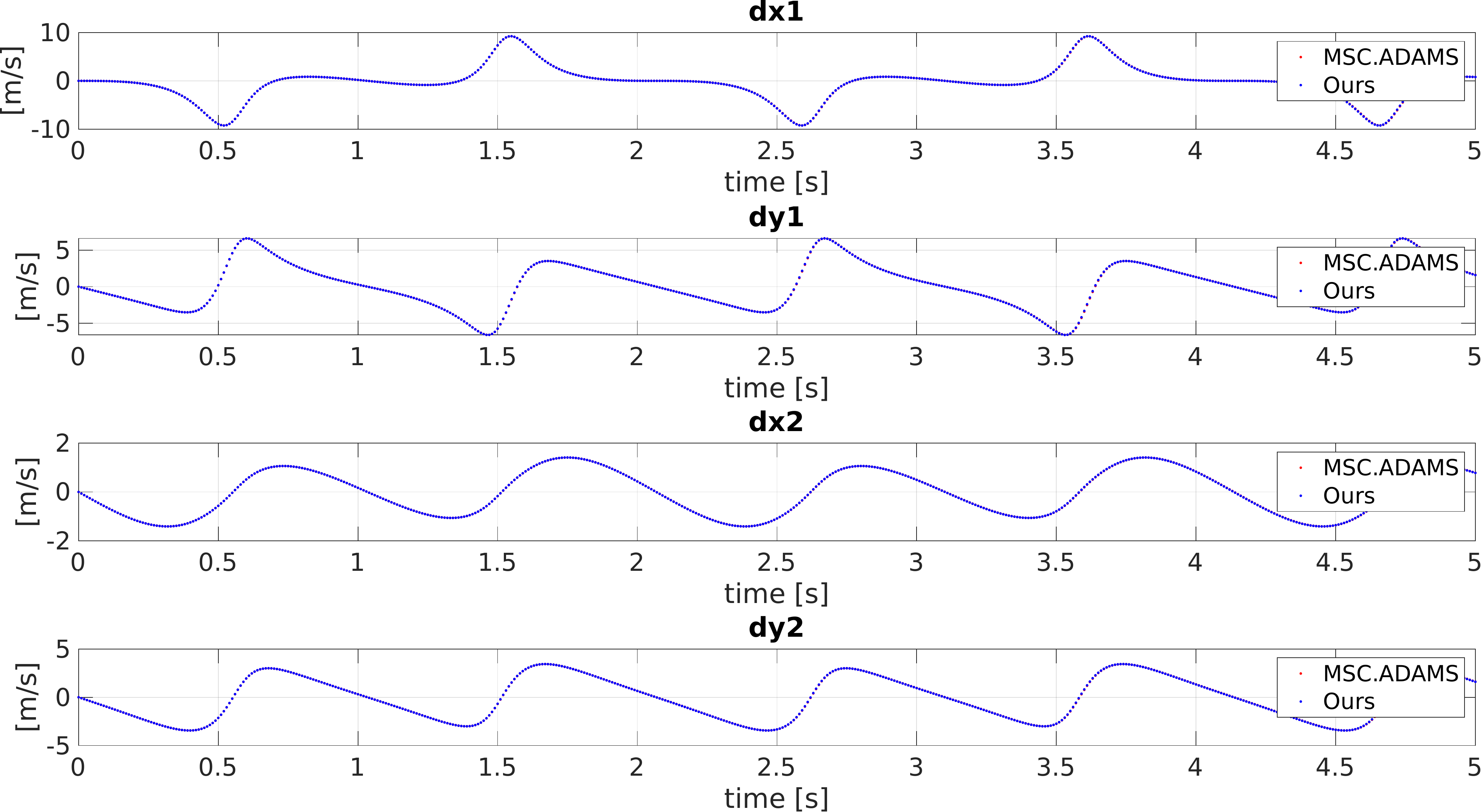}
	\caption{Comparison of $\dq$ series.}
	\end{subfigure}
\\
	\begin{subfigure}[t]{0.46\textwidth}
	\includegraphics[width=1\textwidth]{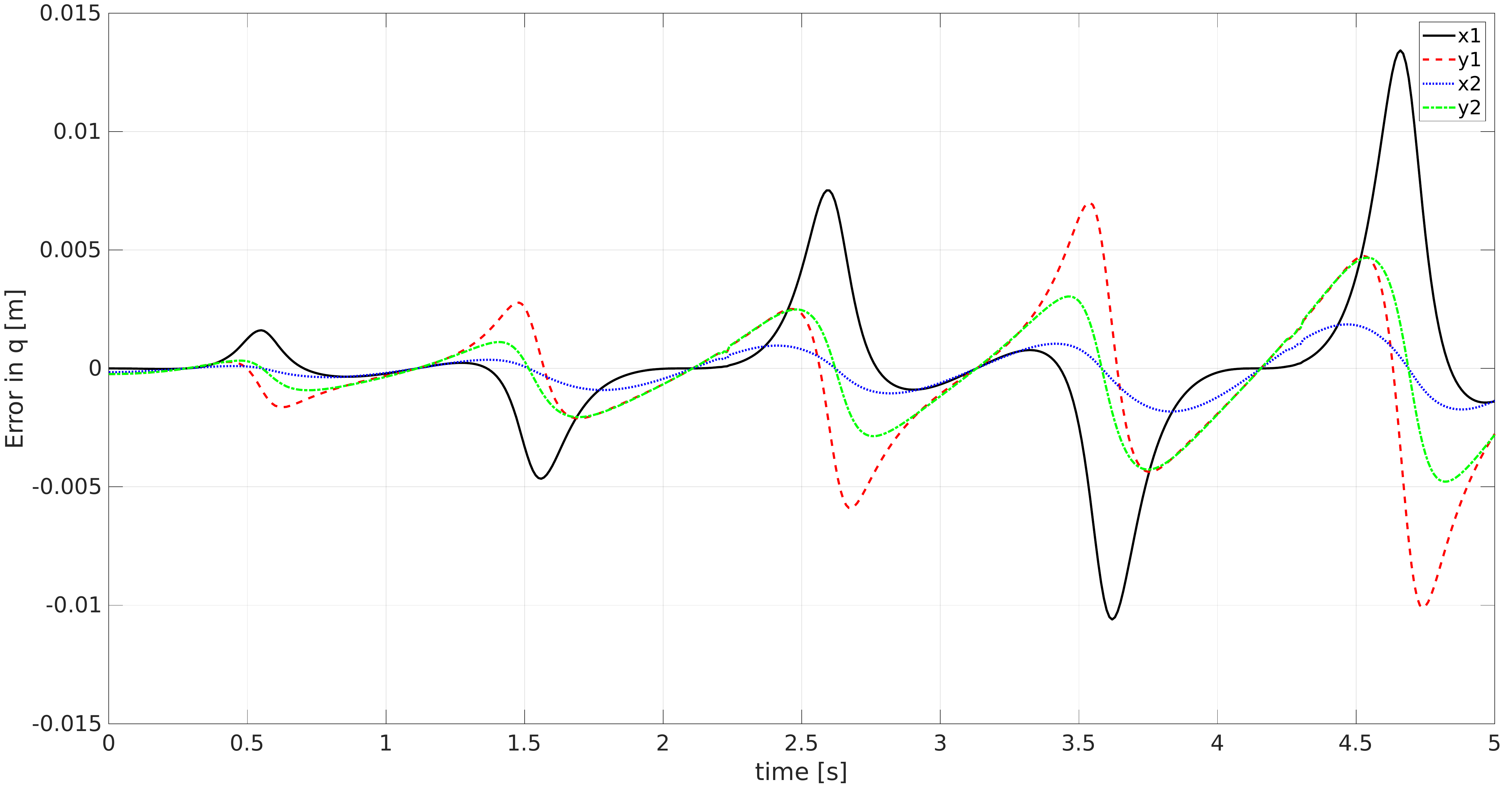}
	\caption{Error in $\q$.}
	\end{subfigure}
	\hspace*{0.5cm}
	\begin{subfigure}[t]{0.46\textwidth}
	\includegraphics[width=1\textwidth]{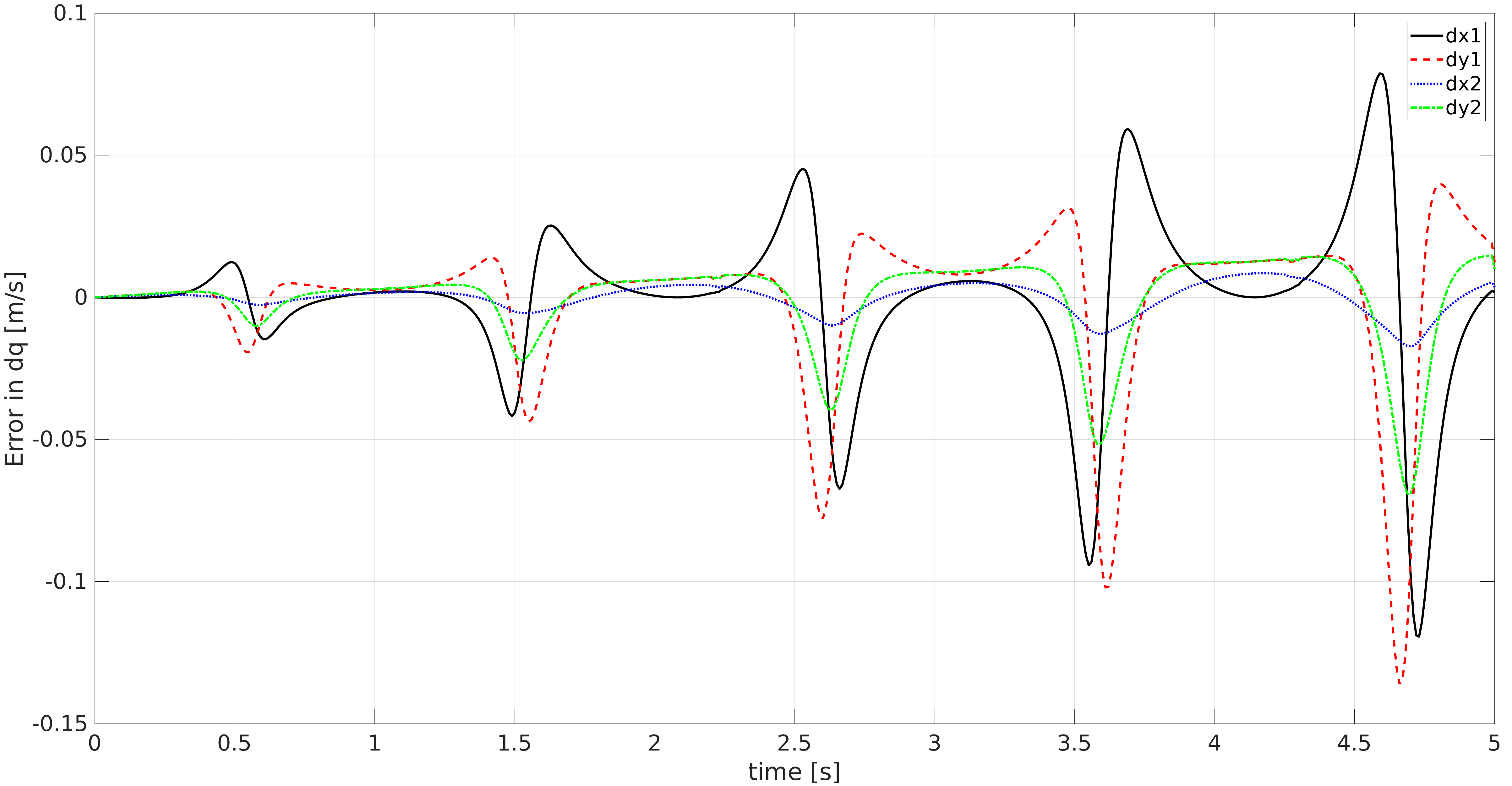}
	\caption{Error in $\dq$.}
	\end{subfigure}
% --
	\caption{Simulation-based validation of the proposed FG approach to forward dynamics for a four-bar linkage under the effects of gravity. Generalized coordinates (a) and velocities (b) are shown for our approach (using a fixed-lag smoother with $N_w=2$, time step of 1 ms) and for MSC.ADAMS as ground truth. (c) and (d) show the error as the difference between ground truth and the output of the proposed FG-based method. Overall RMSE for this case are $\approx 0.0024$ m and $\approx 0.021$ m/s for $\q$ and $\dq$, respectively. See discussion in section~\ref{sect:exp.forward}.}
	\label{fig:exp.comparison1}
\end{figure*}

% --------------------------------------------
\subsection{Forward dynamic simulation}
\label{sect:exp.forward}
% --------------------------------------------
Free motion of the four-bar mechanism under the sole effect of gravity ($g=9.8~ \text{m}/\text{s}^2$) has been simulated,
with an initial crank angle $\theta$ (refer to \FIG{fig:4bar}(a)) of zero and null initial velocities.

The FG in \FIG{fig:fg.dep.forwardDyn} using dependent coordinates has been implemented, and then solved numerically in an incremental fashion
using a fixed-lag smoother, when only the factors of the last $N_w$ time steps (the \emph{window length}) up to time $t$ are considered, with discrete time $t$ ranging from $0$ up to $5$ seconds with steps of $dt=1~\text{ms}$.
A Levenberg-Marquardt nonlinear iterative solver algorithm has been selected for these runs, with a maximum number of iterations of 15.
Note, however, that virtually all calls to the optimizer required only 3 to 5 iterations until convergence.

\FIG{fig:exp.comparison1}(a)--(b) show the time history of the dependent generalized coordinates of points $P_1$ and $P_2$, and their corresponding velocities, respectively,
obtained from both, the proposed FG-based approach
and from MSC Adams.
Both series accurately coincide, hence a more detailed
comparison is in order in \FIG{fig:exp.comparison1}(c)--(d),
where the differences in $\q$ and $\dq$ are reported between the proposed FG-based method and the ground truth.
Similar good agreement (omitted for brevity's sake) is obtained when adopting
the independent coordinates FG, as explained in \FIG{fig:fg.indep.forwardDyn},
thus verifying our two proposed schemes.
The noise models used in all factors involved are summarized in
Table~\ref{tb:fwrd.model.noises}.
Covariance absolute values are not relevant, only their relative
weights with respect to each other.

In order to quantify the accuracy of both schemes (dependent and independent-based FGs) and to explore the effect of the fixed-lag window length parameter $N_w$, the root mean square error (RMSE) of the overall $\q$ trajectories are shown in Table~\ref{tb:rmse.4bars} for window size varying from $2$ to $10$ time steps.
Two main conclusions can be drawn from these data:
(a) the dependent-coordinates scheme seems to provide a higher accuracy,
and (b) using window lengths larger than a few (3--4) time steps
do not further increase the accuracy of the estimation.
This latter point can be explained by the fact that the present
simulations focus on forward dynamics only,
not estimation from \emph{noisy} sensor measurements,
a case where it should be expected that larger windows lead to better results.
This topic, however, falls out of the scope of the present paper and it should
be studied in the future.

\begin{table}[ht]
\centering
  \begin{tabular}{|c|c|c|}
  \hline
  $N_w$ & DC FG  & IC
  FG \\
  $(time steps)$ & RMSE (m)& RMSE (m) \\
  \hline
  2 & 0.002361 & 0.023834 \\
  3 & 0.002356 & 0.019645 \\
  4 & 0.002352 & 0.020019 \\
  5 & 0.002348 & 0.016647 \\
  6 & 0.002344 & 0.016677 \\
  7 & 0.002341 & 0.016677 \\
  8 & 0.002338 & 0.016677 \\
  9 & 0.002335 & 0.017056 \\
 10 & 0.002332 & 0.016677 \\
  \hline
  \end{tabular}
  \caption{RMSE for $\q$ trajectories estimated by the dependent coordinates (DC) and independent coordinates (IC) factor graphs (FG) for the simulation described in the text, with respect to ground-truth, for different window lengths of the fixed-lag smoother estimator.}
  \label{tb:rmse.4bars}
\end{table}

\begin{table}[ht]
\centering
  \begin{tabular}{|c|c|}
  \hline
  Factor & Noise model \\
  \hline
   $f_{prior}(\q_0)$ & $\mathbf{0}_{n \times n}$ \\
 $f_{prior}(\q_0)$ & $diag(\{10^{-3} \text{ if } q(i)\in \z,~1 \text{ otherwise} \}$ \\
 $f_{ti}$ & $10^{-2} ~ \mathbf{I}_{n}$ \\
 $f^d_{dyn},f^i_{dyn}$ & $10^{-4} ~ \mathbf{I}_{n}$ \\
 $f^d_{pc}, f^d_{vc}$ & $\mathbf{0}_{m \times m}$ \\
 $f^i_{pc},f^i_{vc}$ & $\mathbf{0}_{(m+d) \times (m+d)}$ \\
 $f_{eq}$ & $10^{2} ~ \mathbf{I}_{n}$ \\
  \hline
  \end{tabular}
  \caption{Noise models (covariance matrices) employed in the forward dynamics simulations. Most covariances represent isotropic models
  (those including the identity matrix $\mathbf{I}$, or $\mathbf{0}$ for perfectly-known values), with
  only a few of them having different weights for different
  coordinates (those defined with a $diag(\cdots)$, a diagonal matrix with the given diagonal values).}
  \label{tb:fwrd.model.noises}
\end{table}

% --------------------------------------------
\subsection{Inverse dynamic simulation}
\label{sect:exp.inverse}
% --------------------------------------------

Next, the FG proposed in \FIG{fig:fg.inverseDyn}
for inverse dynamic calculation is put at test with the
same four-bar mechanism described above.
Here, the crank angle $\theta$ trajectory over time is specified
as an arbitrary smooth curve, and the goal is to solve for
the unknown torque at the crank (point $A$) that would be required
to balance gravity and inertial forces as accurately following the
prescribed path.

In this case, we use a Levenberg-Marquardt nonlinear solver
on the FG in \FIG{fig:fg.inverseDyn} in batch mode, as opposite to filtering or fixed-lag smoother mode.
However, the position-problem is strongly nonlinear, preventing solvers to find the global optimal
of the whole FG in \FIG{fig:fg.inverseDyn} for arbitrary initial values for all variables. 
Instead, we propose attacking the inverse dynamics FG in the following four stages (running the batch optimizer once after each stage), which ensure that the global optimum is always easy to find by the nonlinear solver:

\begin{itemize}
\item Stage 1: Include $\q$ variables, prior factors for the desired trajectory ($f_{prior}$), position-constraint factors ($f^d_{pc}$), and soft equality factors ($f_{eq}$).
\item Stage 2: Add variables $\dq$, $\ddq$, and numerical integration factors $f_{ti}$.
\item Stage 3: Add velocity-constraint factors ($f^d_{vc}$).
\item Stage 4: Add generalized force variables $\Q$ and inverse dynamics factors ($f^d_{idyn}$).
\end{itemize}

\begin{figure*}
	\centering
	\begin{subfigure}{0.8\textwidth}
		\includegraphics[width=1\textwidth]{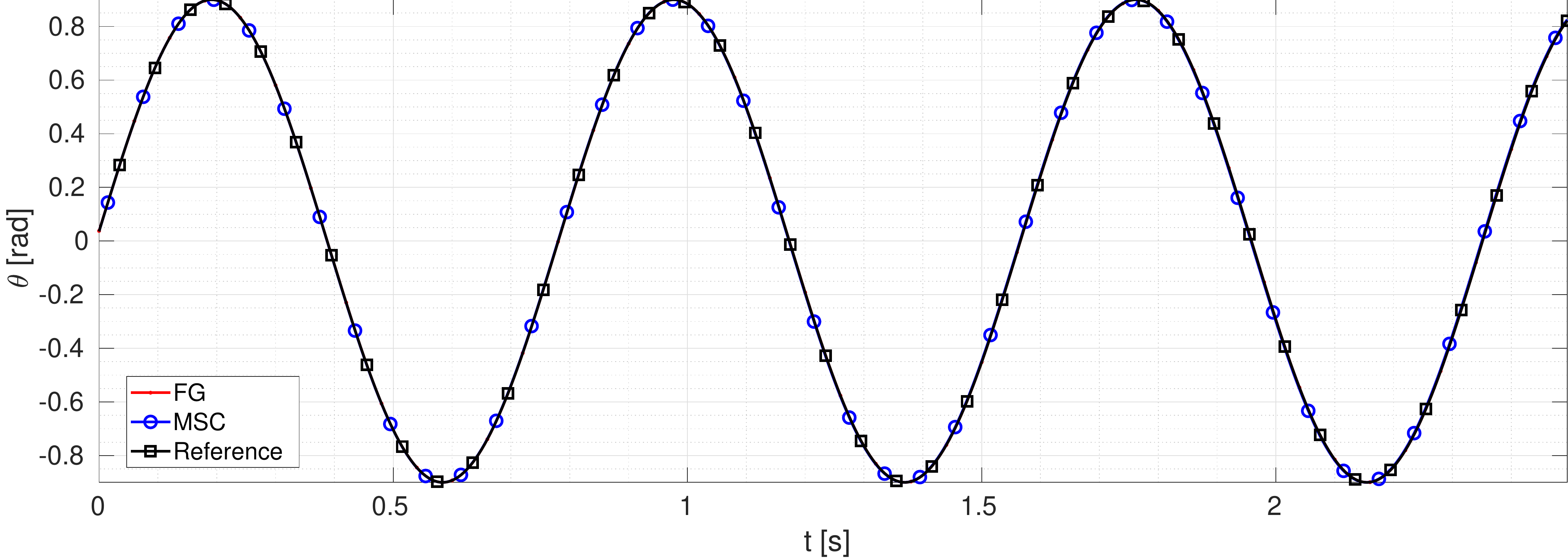}
		\caption{Crank angle reference and simulated trajectories.}
	\end{subfigure}
	\\
	\begin{subfigure}{0.8\textwidth}
	\includegraphics[width=1\textwidth]{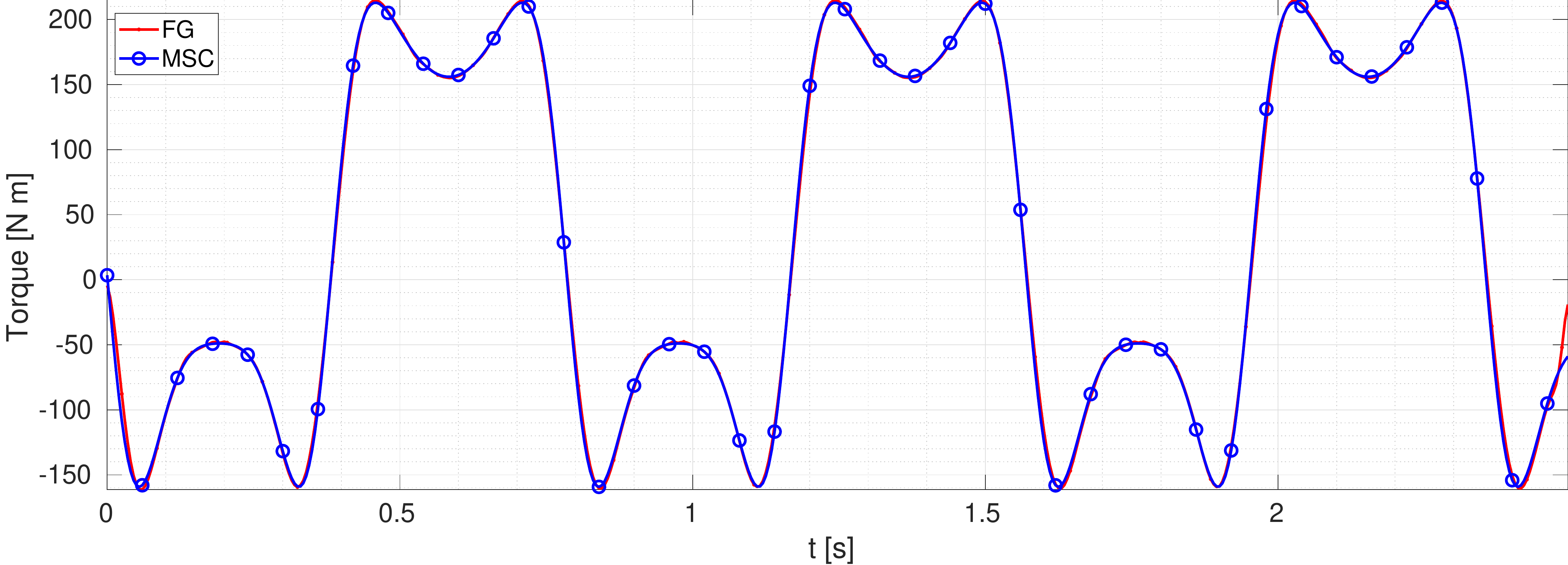}
	\caption{Computed actuator torque.}
	\end{subfigure}
	\\
	\begin{subfigure}{0.8\textwidth}
	\includegraphics[width=1\textwidth]{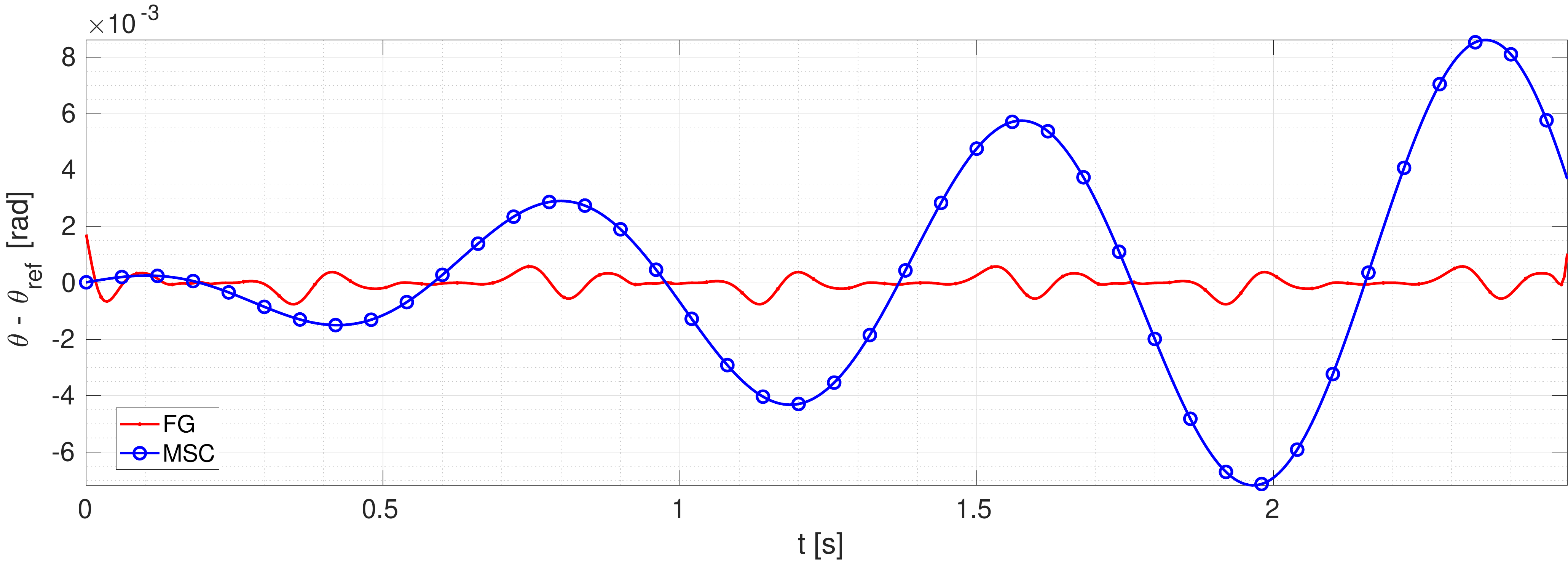}
	\caption{Tracking error of the crank angle.}
	\end{subfigure}
	% --
	\caption{Evaluation of the inverse dynamics test with a four-bar mechanism controlled with a torque at the input crank. Refer to discussion in section~\ref{sect:exp.inverse}.}
	\label{fig:exp.inv.dyn}
\end{figure*}

Numerical results for this approach are shown in \FIG{fig:exp.inv.dyn}, where the trajectories prescribed and actually-followed by the mechanism crank are depicted in \FIG{fig:exp.inv.dyn}(a) for both the proposed approach and the commercial software MSC ADAMS. Both methods successfully solve the inverse dynamics problem by finding the torque sequence in \FIG{fig:exp.inv.dyn}(b).
In this case, our method leads to an excellent stable following error (less than 1 milliradian), whereas MSC's error seems to slowly grow over time, as shown in \FIG{fig:exp.inv.dyn}(c).

% ==================== Conclusions =====================
\section{Conclusions and future works}
\label{sect:Conclusions}

This work has settled the theoretical bases for formulating kinematics and dynamics problems from computational mechanics in the form of factor graphs, and
demonstrated the validity of our reference open-source implementation.
The results showed the practical feasibility of the proposed approach
and its accuracy in comparison with a commercial MBS software based
on classical approaches.
One additional advantage of our formalism
lies at its excellent suitability to design state observers
and parameter identification systems
for arbitrary mechanical systems, ranging from robots to vehicles. This aspect will be part of future works.

%\section*{Acknowledgments}
%This work has been partially funded by XXX ...

% ====================== Appendix ======================
%\renewcommand{\theequation}{A-\arabic{equation}}
%\setcounter{equation}{0}  % reset counter
%\appendix

\section*{Appendix I. Block matrix inversion lemma}
\label{sect:Appendix.I.BlockMatrix}

This lemma implies that:

\begin{equation}
\begin{bmatrix}
\mathbf{A} && \mathbf{B} \\
\mathbf{C} && \mathbf{D}
\end{bmatrix}^{-1} =
\begin{bmatrix}
\mathbf{A}^{-1}+\mathbf{A}^{-1}\mathbf{B}\left(\mathbf{D}-\mathbf{C}\mathbf{A}^{-1}\mathbf{B}\right)^{-1}\mathbf{C}\mathbf{A}^{-1} &&
-\mathbf{A}^{-1}\mathbf{B}\left(\mathbf{D}-\mathbf{C}\mathbf{A}^{-1}\mathbf{B}\right)^{-1} \\
-\left(\mathbf{D}-\mathbf{C}\mathbf{A}^{-1}\mathbf{B}\right)^{-1}\mathbf{C}\mathbf{A}^{-1} &&
\left(\mathbf{D}-\mathbf{C}\mathbf{A}^{-1}\mathbf{B}\right)^{-1}
\end{bmatrix}
\label{ApxEq: BlockMatInv}
\end{equation}

%
% JL: Uncomment and move as a footnote where explicitly used:
%
%Below some matrices properties exploited in this paper. \\
%\textbf{(Moore-Penrose) Pseudoinverse Matrix Properties $\mathbf{A}^{+}$}
%\begin{equation}
%\mathbf{A}\mathbf{A}^{+} = \mathbf{I} \neq \mathbf{A}^{+}\mathbf{A}
%\label{ApxEq: PseudoInv1}
%\end{equation}
%
%
% Derivative of an Inverse Matrix:
%
%\begin{equation}
%
%\label{ApxEq:InvDerivative}
%\end{equation}
%

\section*{Appendix II. Tensor notation definitions and useful expressions}

\emph{Order of a tensor}: Number of indices to index all its components, i.e. Scalars, vectors, and matrices are 0th, 1st, and 2nd-order tensors, respectively.
		
\emph{Einstein convention}: Superscript indices imply reading elements up-down, subscript indices imply reading left-right. Summation works over the repeated index or indices. Example with a common matrix product:
		
\begin{eqnarray}
	\mathbf{C}=\mathbf{AB} \Rightarrow
	\left\{
	\begin{array}{lc}
		\text{Conventional notation:}  &  C_{ik} = \sum_j A_{ij} B_{jk} \\
		\text{Einstein convention:} &  C^i{}_k = A^i{}_j B^j{}_k
	\end{array}
	\right.
\end{eqnarray}

\emph{Transpose rule}: We have that $A^i{}_j = (A^\top)^j{}_i$.

\emph{Jacobian}: $\frac{\partial \mathbf{x}}{\partial \mathbf{y}}$, assuming $\mathbf{x}$ and $\mathbf{y}$ are a row and a column vector, respectively, becomes:
\begin{eqnarray}
	\frac{\partial \mathbf{x}}{\partial \mathbf{y}} = \frac{\partial x_i}{\partial y_j} = \Delta_i{}^j
\end{eqnarray}

\emph{The standard derivative chain rule}: Applied to vector variables $\mathbf{x}$ and $\mathbf{z}$, using an intermediary
variable $\mathbf{y}$, can be put in tensor notation as:

\begin{eqnarray}
	\frac{\partial \mathbf{x}}{\partial \mathbf{y}}
	= \left. \frac{\partial x_i}{\partial z_k} \right|_{n_i \times n_k}
	= \frac{\partial x_i}{\partial y_j} \frac{\partial y_j}{\partial z_k}
	= \underbrace{ \Delta^i{}_j }_{n_i \times n_j} \underbrace{ \Delta^j{}_k }_{n_j \times n_k}
	\equiv \sum_j  \frac{\partial x_i}{\partial y_j} \frac{\partial y_j}{\partial z_k}
\end{eqnarray}

\emph{Chain rule for derivatives of matrices with respect to a vector}: It gives us the third-order tensor:

\begin{eqnarray}
	\frac{\partial \mathbf{A}}{\partial \mathbf{z}}
	&=& \underbrace{  \frac{\partial A^i{}_j}{\partial z_k} }_{n_i \times n_j \times n_k}
	= \underbrace{ \frac{\partial A^i{}_j}{\partial y_l} }_{n_i \times n_j \times n_l}
	\underbrace{ \frac{\partial y_l}{\partial z_k} }_{n_l \times n_k}
\end{eqnarray}

\noindent where it becomes clear that we must sum over the $l$ indices, that is, converting the Einstein notation above into a conventional summation:

\begin{eqnarray}
	\frac{\partial \mathbf{A}}{\partial \mathbf{z}}
	&=&
	\sum_l \underbrace{ \frac{\partial \mathbf{A}}{\partial y_l} }_{n_i \times n_j \times n_l}
	\underbrace{ \frac{\partial y_l}{\partial \mathbf{z} } }_{n_l \times n_k}
\end{eqnarray}

%\emph{Particular case}: A case we often find in computational Mechanics is the derivative
%of a matrix with respect a scalar variable, using a vector as intermediary variable for the chain rule.
%A third-order tensor appears in the expressions but the final result is a regular 2-D matrix:
%
%\begin{eqnarray}
%	\frac{\partial \mathbf{A}}{\partial t}
%	&=& \underbrace{  \frac{\partial A^i{}_j}{\partial t} }_{n_i \times n_j}
%	= \underbrace{ \frac{\partial A^i{}_j}{\partial y_l} }_{n_i \times n_j \times n_l}
%	\underbrace{ \frac{\partial y_l}{\partial t} }_{n_l \times 1} \\
%	&=&
%	\sum_l \underbrace{ \frac{\partial \mathbf{A}}{\partial y_l} }_{n_i \times n_j}
%	\underbrace{ \frac{\partial y_l}{\partial t } }_{n_l \times 1}
%\end{eqnarray}

\emph{Product of a third-order tensor and a vector}: Given a
third-order tensor $\mathbf{T} = T^i_j{~}^{k}$
and a vector $\mathbf{v} = v_k$,
its product reads:
\begin{equation}
\label{ApsEq:innVecTens}
\mathbf{T} \mathbf{v}
=
T^i_j{~}^{k} ~ v_k
=
\sum_k
T^i_j{}^{k} v_k
%\exists \hspace*{0.1cm} \mathbf{M}_{ij}\in\mathbb{R}^{i\times j} \Spp \text{such that} \Spp \mathbf{M}_{ij} = \mathbf{v}_k^\top \cdot \mathbf{T}_{kij}
\end{equation}
%\noindent and, moreover, $\mathbf{v}_k^\top \cdot \mathbf{T}_{kij} \ne \mathbf{T}_{ijk}\cdot \mathbf{v}_k$.

% ==============================================================================================
% APPENDIX III
% ==============================================================================================
\section*{Appendix III. Equations for planar multibody systems}
\label{sect:Appendix.Eqs.MBD.elements}

% Collect (maybe in a single figure) analysed constraints images.

In the following, we summarize the equations implemented in our open-source library
to model typical mechanical constraints of planar mechanisms.
These equations are the building blocks of the errors and Jacobian matrices associated with
the factors described in sections~\ref{sect:factor.pos.constr.dep}-\ref{sect:factor.vel.constr.dep}.
An open-source implementation of all these equations is provided in
the C++ toolkit mentioned in section~\ref{sect:Experiments}.

Since each constraint contributes only a few scalar components to $\PHI$ (typically only one for planar mechanisms), the following uses the superscript $\PHI^k$ to reflect the $k$-th row in the corresponding vector or matrix, with $k$ being arbitrarily determined by the order in which constraints are enumerated when defining a mechanism.
Furthermore, for the sake of notation clarity, $\q$ should be understood in the following sets of equations as the subset of the actual $\q$ that contains
only those generalized coordinates that are directly involved in the constraint, i.e. the provided Jacobians should be expanded with columns of zeros as needed at the position of non-relevant coordinates.

In addition to the specific formulas for each constraint,
the following terms appear in factor Jacobians but are
generic so they apply to all constraints:

\begin{subeqnarray}
\label{eq:apx.common.mbs}
\dot{\PHI}_\q^k(\q) &=& \dfrac{d\Phiq^k(\q)}{dt} \hspace*{0.2cm }= \Sp
			\dfrac{\partial \Phiq^k}{\partial \q}\dq + \cancelto{0}{\PHI_t} \hspace*{0.2cm }= \Sp
			\Phiqq^k\dq
\slabel{eq:apx.common.mbs.a}
\\
\frac{\partial \dot{\PHI}^k_{\q}(\q)}{\partial \dq}  &=&  \Phiqq^k \quad \text{(From \EQ{eq:apx.common.mbs.a})}
\\
\dfrac{\partial \lp \Phiq^k \lp \q \rp \dq \rp}{\partial \q}  &=& \Phiqq^k~ \dq \hspace*{0.2cm } = \dot{\PHI}_\q^k(\q) \\
\dfrac{\partial \lp \Phiq^k \lp \q \rp \dq \rp}{\partial \dq} &=& \Phiq^k
\end{subeqnarray}

\subsection*{Constant distance}

This constraint is typically used to model rigid solids.
Given the coordinates of any pair of points $i$ and $j$
of a rigid body, such that $(x_i, y_i) \in \q$ and $(x_j,y_j) \in \q$, this constraint ensures a fixed distance $L$ between them.
Each such constraint contributes with the following rows
in the different vectors and matrices defining the multibody system:
	
\begin{subeqnarray}
\PHI^k(\q) &=& \lp x_j-x_i\rp^2 + \lp y_j-y_i\rp^2 - L^2 = 0
\\
% -----
\Phiq^k(\q) &=& \left[
2\lp x_i-x_j\rp \Spp 2\lp y_i-y_j\rp \Spp 2\lp x_j-x_i\rp \Spp 2\lp y_j-y_i\rp \right]
\\
% -----
\dot{\PHI}_\q^k(\q) &=& \left[
	2\lp \dot{x}_i-\dot{x}_j\rp \Spp 2\lp \dot{y}_i-\dot{y}_j\rp \Spp 2\lp \dot{x}_j-\dot{x}_i\rp \Spp 2\lp \dot{y}_j-\dot{y}_i\rp \right]
% -----
\\
\Phiqq^k(\q) = \dfrac{\partial \Phiq^k}{\partial \q} &=&
\left[ \dfrac{\partial \Phiq^k}{\partial x_1} \Spp \dfrac{\partial \Phiq^k}{\partial y_1} \Spp \dfrac{\partial \Phiq^k}{\partial x_2} \Spp \dfrac{\partial \Phiq^k}{\partial y_2}
\right] =
  \begin{blockarray}{ccccc}
x_i & y_i & x_j & y_j \\
\begin{block}{[cccc]c}
  2 & 0 & -2 & 0 & \Phiq_1^k \\
  0 & 2 & 0 & -2 & \Phiq_2^k \\
  -2 & 0 & 2 & 0 & \Phiq_3^k \\
  0 & -2 & 0 & 2 & \Phiq_4^k \\
\end{block}
\end{blockarray}
% -----
\\
\dotPhiqq^k(\q) &=& \mathbf{0}_{4 \times 4}
\\
\Phiqq^k(\q) \ddq &=& \left[
	2\lp \ddot{x}_i-\ddot{x}_j\rp \Spp 2\lp \ddot{y}_i-\ddot{y}_j\rp \Spp 2\lp \ddot{x}_j-\ddot{x}_i\rp \Spp 2\lp \ddot{y}_j-\ddot{y}_i\rp \right]
\end{subeqnarray}

\subsection*{Fixed pinned slider constraint}

This constraints model a point $P(x,y)$ enforced to move along a line defined by two fixed points $A(x_A, y_A)$ and $B(x_B, y_B)$.
By exploiting the properties of the similar triangles, one obtains:

\begin{subeqnarray}
%\label{eq:FixSlider}
\PHI^k(\q) &=& (x_B - x_A)\lp y-y_A\rp- (y_B - y_A)\lp x-x_A\rp \\
% -----
\Phiq^k(\q) &=& \left[-(y_B - y_A) \hspace*{0.35cm} (x_B - x_A) \right] \\
% -----
%\dot{\PHI}_\q^k(\q)\dq &=& \Delta_x \dot{y} - \Delta_y \dot{x} \\
% -----
%\Phiqq^k(\q) &=& \mathbf{0} \\
% -----
\dot{\PHI}_\q^k(\q) &=& \mathbf{0}_{1 \times 2}
% -----
\\
\Phiqq^k(\q) &=& \mathbf{0}_{2 \times 2}
\end{subeqnarray}
	
\subsection*{Mobile pinned slider constraint}

A more general version of the former constraint,
where the point P$(x,y)$ is a now moving
along a line defined by two \emph{mobile} points A$(x_i,y_i)$ and B$(x_j,y_j)$.
In this case:

\begin{subeqnarray}
%\label{eq:MobileSlider}
% JL: **This particular line renders better without \lp \rp**
\PHI^k(\q) &=& ( x_j-x_i) ( y-y_i)-(y_j-y_i)(x-x_i) \\
% -----
\Phiq^k(\q) &=& \left[y_i-y_j \Spp x_j-x_i \Spp y_j-y \Spp x-x_j \Spp y-y_i \Spp x_i-x \right] \\
% -----
\dot{\PHI}_\q^k(\q)  &=& \left[\dot{y}_i-\dot{y}_j \Spp \dot{x}_j-\dot{x}_i \Spp \dot{y}_j-\dot{y} \Spp \dot{x}-\dot{x}_j \Spp \dot{y}-\dot{y}_i \Spp \dot{x}_i-\dot{x} \right]
% -----
\\
\Phiqq^k(\q) &=&
\begin{blockarray}{*{7}{c}}
x & y & x_i & y_i & x_j & y_j \\
\begin{block}{[cccccc]c}
0 & 0 & 0 & 1 & 0 & -1 & \Phiq_1^k \\
0 & 0 & -1 & 0 & 1 & 0 & \Phiq_2^k \\
0 & -1 & 0 & 0 & 0 & 1 & \Phiq_3^k \\
1 & 0 & 0 & 0 & -1 & 0 & \Phiq_4^k \\
0 & 1 & 0 & -1 & 0 & 0 & \Phiq_5^k \\
-1 & 0 & 1 & 0 & 0 & 0 & \Phiq_6^k \\
\end{block}
\end{blockarray} \\
% -----
\dotPhiqq^k(\q) &=& \mathbf{0}_{6 \times 6}
\\
\Phiqq^k(\q) \ddq &=& \left[
  \ddot{y}_i-\ddot{y}_j \Spp \ddot{x}_j-\ddot{x}_i \Spp \ddot{y}_j-\ddot{y} \Spp \ddot{x}-\ddot{x}_j \Spp \ddot{y}-\ddot{y}_i \Spp \ddot{x}_i-\ddot{x} \right]
\end{subeqnarray}
	
%
% JLBC: I'm not 100 % sure of this one! will comment it out for now.
%
%\subsection*{Rigid slider constraint}
%
%This constraint allows the translation of a rigid body
%along
%with respect to a specific direction (A-B in the figure). The angle $\theta$ between the horizontal and the line C-D (of length $L$) is constant. Since $L$ is known, this constraint forces the equality of \EQ{eq:RigidSlider.a}, where the points C$(x_i,y_i)$ and D$(x_j,y_j)$ belong to $\q$.
%
%\begin{subeqnarray}
%\label{eq:RigidSlider}
%\slabel{eq:RigidSlider.a}
%\PHI^k(\q) &=& \lp x_j-x_i \rp + \lp y_j-y_i \rp - L \lp \sin\theta + \cos\theta \rp \\
%% -----
%\Phiq^k(\q) &=& \left[-1 \Spp -1 \hspace*{0.35cm} 1 \hspace*{0.35cm} 1 \right] \\
%% -----
%\dot{\PHI}_\q^k(\q) &=& \mathbf{0}_{1 \times 4} \\
%% -----
%\Phiqq^k(\q) &=& \dotPhiqq^k(\q) = \mathbf{0}_{4 \times 4}
%\end{subeqnarray}
%
%This constraint includes two particular cases:
%\begin{itemize}
%	\item \textit{Vertical rigid slider}
%	\begin{eqnarray}
%	&&y_1=y_2, \Spp \sin\theta=1 \Spp \text{and} \Spp \cos\theta=0 \nonumber \\
%	\PHI^k(\q) &=& \lp y_2-y_1 \rp - L \\
%	\Phiq^k(\q) &=&
%	\left[0 \Spp -1 \hspace*{0.35cm} 0 \hspace*{0.35cm} 1 \right] \nonumber
%	\end{eqnarray}
%	\item \textit{Horizontal rigid slider}
%	\begin{eqnarray}
%	&&x_1=x_2, \Spp \sin\theta=0 \Spp \text{and} \Spp \cos\theta=1 \nonumber \\
%	\PHI^k(\q) &=& \lp x_2-x_1 \rp - L \\
%	\Phiq^k(\q) &=&
%	\left[-1 \Spp 0 \hspace*{0.35cm} 1 \hspace*{0.35cm} 0 \right] \nonumber
%	\end{eqnarray}
%\end{itemize}

\subsection*{Absolute orientation of planar link}

Another special kind of constraints is that one needed when defining relative coordinates, such as the absolute angle of a link, e.g. a mechanism input crank, or a vehicle wheel position.
The relative angle discussed in this section is called \emph{absolute} since it is defined as the angle $\theta$ between a body comprising a pair of points $(x_i,y_i)$--$(x_j,y_j)$, at a fixed distance $L$, and the horizontal axis of the global frame of reference.

For this constraint, with the ordering of relevant coordinates being $\q=[x_i ~ y_i ~ x_j ~ y_j ~ \theta]^\top$, two sets of equations exist to avoid degeneracy.
The first ones are used when $|\sin(\theta)|>1/\sqrt{2}$:

\begin{subeqnarray}
\PHI^k(\q) &=& x_j-x_i - L \cos(\theta) = 0
\\
\Phiq^k(\q) &=& \left[
-1 \Sp 0 \Sp 1 \Sp 0 \Sp L \sin(\theta) \right]
\\
\dot{\PHI}_\q^k(\q) &=& \Sp \left[
0 \Sp 0 \Sp 0 \Sp 0 \Sp L \dot{\theta} \cos(\theta)  \right]
\\
% -----
\Phiqq^k(\q) &=&
\begin{blockarray}{*{6}{c}}
x_i & y_i & x_j & y_j & \theta \\
\begin{block}{[ccccc]c}
0 & 0 & 0 & 0 & 0  & \Phiq_1^k \\
0 & 0 & 0 & 0 & 0 & \Phiq_2^k \\
0 & 0 & 0 & 0 & 0 & \Phiq_3^k \\
0 & 0 & 0 & 0 & 0 & \Phiq_4^k \\
0 & 0 & 0 & 0 & L \cos \lp \theta \rp & \Phiq_5^k \\
\end{block}
\end{blockarray} \\
% -----
\dotPhiqq^k(\q) &=&
\begin{blockarray}{*{6}{c}}
x_i & y_i & x_j & y_j & \theta \\
\begin{block}{[ccccc]c}
0 & 0 & 0 & 0 & 0  & \dot{\PHI}_{\q 1}^k \\
0 & 0 & 0 & 0 & 0 & \dot{\PHI}_{\q 2}^k \\
0 & 0 & 0 & 0 & 0 & \dot{\PHI}_{\q 3}^k \\
0 & 0 & 0 & 0 & 0 & \dot{\PHI}_{\q 4}^k \\
0 & 0 & 0 & 0 & - L \dot{\theta} \sin \lp \theta \rp & \dot{\PHI}_{\q 5}^k \\
\end{block}
\end{blockarray}
\\
\Phiqq^k(\q) \ddq &=& \left[
  0 \Spp 0 \Spp 0 \Spp 0 \Spp L \ddot{\theta} \cos \lp \theta \rp \right]
\\
\dotPhiqq^k(\q) \dq &=& \left[
  0 \Spp 0 \Spp 0 \Spp 0 \Spp - L \dot{\theta}^2 \sin \lp \theta \rp \right]
\end{subeqnarray}

\noindent and the following alternative ones are used otherwise:

\begin{subeqnarray}
\PHI^k(\q) &=& y_j-y_i - L \sin(\theta) = 0
\\
\Phiq^k(\q) &=& \left[
0 \Sp -1 \Sp 0 \Sp 1 \Sp - L \cos(\theta) \right]
\\
\dot{\PHI}_\q^k(\q) &=& \Sp \left[
0 \Sp 0 \Sp 0 \Sp 0 \Sp L \dot{\theta} \sin(\theta) \right]
\\
% -----
\Phiqq^k(\q) &=&
\begin{blockarray}{*{6}{c}}
x_i & y_i & x_j & y_j & \theta \\
\begin{block}{[ccccc]c}
0 & 0 & 0 & 0 & 0  & \Phiq_1^k \\
0 & 0 & 0 & 0 & 0 & \Phiq_2^k \\
0 & 0 & 0 & 0 & 0 & \Phiq_3^k \\
0 & 0 & 0 & 0 & 0 & \Phiq_4^k \\
0 & 0 & 0 & 0 & L \sin \lp \theta \rp & \Phiq_5^k \\
\end{block}
\end{blockarray} \\
% -----
\dotPhiqq^k(\q) &=&
\begin{blockarray}{*{6}{c}}
x_i & y_i & x_j & y_j & \theta \\
\begin{block}{[ccccc]c}
0 & 0 & 0 & 0 & 0  & \dot{\PHI}_{\q 1}^k \\
0 & 0 & 0 & 0 & 0 & \dot{\PHI}_{\q 2}^k \\
0 & 0 & 0 & 0 & 0 & \dot{\PHI}_{\q 3}^k \\
0 & 0 & 0 & 0 & 0 & \dot{\PHI}_{\q 4}^k \\
0 & 0 & 0 & 0 & L \dot{\theta} \cos \lp \theta \rp & \dot{\PHI}_{\q 5}^k \\
\end{block}
\end{blockarray}
\\
\Phiqq^k(\q) \ddq &=& \left[
  0 \Spp 0 \Spp 0 \Spp 0 \Spp L \ddot{\theta} \sin \lp \theta \rp \right]
\\
\dotPhiqq^k(\q) \dq &=& \left[
  0 \Spp 0 \Spp 0 \Spp 0 \Spp L \dot{\theta}^2 \cos \lp \theta \rp \right]
\end{subeqnarray}

% ===================== References =====================
%\section*{References}
\bibliographystyle{elsarticle-num}
\bibliography{references}

\end{document}